\documentclass{article} 
\usepackage[preprint]{colm2025_conference}
\usepackage{microtype}
\usepackage{hyperref}
\usepackage{url}
\usepackage{booktabs}
\usepackage{amsmath}
\usepackage{lineno}
\usepackage{subfigure} 
\usepackage{graphicx}    
\usepackage{adjustbox}   
\usepackage{caption} 
\usepackage{subcaption}
\usepackage{float}

\definecolor{darkblue}{rgb}{0, 0, 0.5}
\hypersetup{colorlinks=true, citecolor=darkblue, linkcolor=darkblue, urlcolor=darkblue}

\title{Adversarial Training of Reward Models}

\title{Adversarial Training of Reward Models}

\author{
  \textbf{Alexander Bukharin\textsuperscript{1,2}}, \quad
  \textbf{Haifeng Qian\textsuperscript{1}}, \quad
  \textbf{Shengyang Sun\textsuperscript{1}}, \\
  \textbf{~Adithya Renduchintala\textsuperscript{1}}, \quad
  \textbf{Soumye Singhal\textsuperscript{1}}, \quad
  \textbf{Zhilin Wang\textsuperscript{1}}, \quad
  \\ \textbf{~Oleksii Kuchaiev\textsuperscript{1}},
  \textbf{Olivier Delalleau\textsuperscript{1}}, \quad
  \textbf{Tuo Zhao\textsuperscript{2}} \\
  ~\textsuperscript{1}NVIDIA \quad \textsuperscript{2}Georgia Institute of Technology
}

\begin{document}

\ifcolmsubmission
\linenumbers
\fi

\maketitle

\begin{abstract}
Reward modeling has emerged as a promising approach for the scalable alignment of language models. However, contemporary reward models (RMs) often lack robustness, awarding high rewards to low-quality, out-of-distribution (OOD) samples. This can lead to reward hacking, where policies exploit unintended shortcuts to maximize rewards, undermining alignment. To address this challenge, we introduce Adv-RM, a novel adversarial training framework that automatically identifies adversarial examples — responses that receive high rewards from the target RM but are OOD and of low quality. By leveraging reinforcement learning, Adv-RM trains a policy to generate adversarial examples that reliably expose vulnerabilities in large state-of-the-art reward models such as Nemotron 340B RM. Incorporating these adversarial examples into the reward training process improves the robustness of RMs, mitigating reward hacking and enhancing downstream performance in RLHF. We demonstrate that Adv-RM significantly outperforms conventional RM training, increasing stability and enabling more effective RLHF training in both synthetic and real-data settings. We will open-source all code and data.
\end{abstract}


\section{Introduction}
As large language models (LLMs) become increasingly capable, ensuring their alignment with human values remains a fundamental challenge \citep{amodei2016concrete, askell2021general}. One of the most promising approaches for scalable alignment of LLMs is reward modeling, in which a reward model is trained to predict human feedback \citep{christiano2017deep, leike2018scalable}. These reward models (RMs) are then used as a proxy for human feedback in a variety of places throughout the LLM lifecycle, including data selection \citep{chen2023alpagasus}, Reinforcement Learning from Human Feedback (RLHF, \citealt{ouyang2022training}), and inference time moderation \citep{inan2023llama}. This approach has been very successful, resulting in powerful language models such as GPT-4 and Llama 3 \citep{achiam2023gpt, dubey2024llama}.

Despite their widespread use in LLM alignment, RMs are imperfect proxies for human judgment \citep{hendrycks2021unsolved, gao2023scaling}. In particular, they exhibit poor out-of-distribution (OOD) generalization, often failing to reliably assess prompt-response pairs that diverge from their training distribution \citep{eisenstein2023helping}. OOD generalization is challenging because RMs learn from a limited set of human-annotated examples, which do not cover the full diversity of possible model behaviors, leading to systematic failures when encountering novel model responses. These failures lead to reward hacking—where models trained on RMs optimize reward scores through unintended shortcuts, reducing their actual alignment with human values. Such reward hacking can result in unexpected behaviors and difficult-to-detect misalignment, an issue that becomes even more pernicious as model capabilities increase and evaluation becomes more challenging. 

Building robust RMs to mitigate reward hacking presents two unsolved challenges. The first is measuring a RM's robustness. Unlike image and speech models, where adversarial attacks can serve as robustness metrics \citep{carlini2017towards}, the discrete nature of language makes adversarial attacks on language models challenging and their robustness harder to define \citep{wang2021adversarial}. Reward models introduce another layer of complexity, as their accuracy is measured on preference pairs, requiring adversarial preference pairs rather than a single adversarial input. So far the only known reliable way to quantify robustness of a RM is to measure how long it takes for reward hacking to occur in RLHF. This in itself can be challenging: we can rely on human evaluation to tell us if the policy training has significantly degraded, but this is costly and small misalignments can easily go undetected.

The second problem is how to improve the robustness of a RM.
Prior works show that reward hacking often occurs because the RM gives high reward scores to model responses outside the reward model training distribution \citep{casper2023open, eisenstein2023helping}. Giving a high reward to OOD responses is bad, as they deviate significantly from the expert responses in the RLHF dataset and are therefore likely to be low quality \citep{ho2016generative}. To avoid rewarding such responses, contemporary works have tried a variety of approaches, including regularizing training so that the policy model remains close to its initialization \citep{jaques2019way, stiennon2020learning}, penalizing out-of-distribution (OOD) model responses using uncertainty estimates \citep{eisenstein2023helping, coste2023reward}, and regularizing reward learning \citep{ibarz2018reward, liu2024rrm}. 
However, we find that uncertainty estimates are not reliable indicators of response quality for in-distribution responses, which may limit the effectiveness of such regularization.

In this work we propose a new adversarial training framework, Adv-RM, to address both problems. Adv-RM automatically discovers adversarial responses that the targeted RM gives a high reward to while being OOD and low quality. We find these examples by training an adversarial policy using reinforcement learning to generate responses that maximize discrepancy across different RMs, while receiving high rewards from the target RM. This approach reliably finds adversarial examples for state-of-the-art (SOTA) RMs, including Skywork-Reward-Gemma-2-27B \citep{liu2024skywork}, Llama-3.1-Nemotron-70B-Reward \citep{wang2024helpsteer2}, and Nemotron-4-340B-Reward \citep{adler2024nemotron}, with over 80\% attack success rates. Interestingly, we find that these popular RMs give high reward scores to a variety of low-quality responses, such as responses that have no punctuation or responses with random text in the middle of a response.

Once we find adversarial samples, Adv-RM incorporates them into the RM training procedure. Since the success rate of our attack method is high, we can include adversarial samples in the training data set without additional human annotation cost by simply using an in-distribution response as the chosen response and the adversarial response as the rejected response. We conduct extensive evaluation of these adversarially trained RMs through adversarial attacks, downstream RLHF training in the synthetic setting of \citet{gao2023scaling}, and RLHF training on real datasets \citep{wang2024helpsteer2}. We found that Adv-RM reward models not only are more robust against attacks but also mitigate reward hacking. With these models, RLHF can proceed for three times as many steps as runs with conventional RMs without exhibiting reward hacking, resulting in more performant aligned models.

We summarize our contributions as follows:


$\bullet$ We develop an RL-based approach that successfully generates adversarial samples for SOTA RMs. To the best of our knowledge, this is the first work to do so without human domain knowledge, providing a new means of evaluating RM robustness.


$\bullet$ We use adversarial samples to augment RM training. Through extensive experiments, we find that such adversarial training results in reward models that are more robust to hacking and have better downstream performance in RLHF.

\section{Related Work}
\label{sec:related-work}

\textbf{Mitigating Reward Hacking.} There have been several works that attempt to mitigate reward hacking. Some of the earliest works used reward ensembles, and found that different ensembling approaches can slow (but not prevent) reward hacking \citep{coste2023reward, eisenstein2023helping, zhai2023uncertainty, rame2024warm, yan2024reward}. In our experiments we compare with two simple reward ensemble aggregation approaches, MEAN and MEAN\_MINUS\_STD \citep{eisenstein2023helping}. \citet{chen2024odin} propose a new architecture that disentangles length from human preferences. Recently \citet{liu2024rrm} proposes RRM which augments the RM training data by comparing responses from a prompt with random responses from other prompts, an approach we find to be effective. 

Two works discuss adversarial training of verifiers. \citet{amodei2016concrete} presents the idea of adversarial reward models, a training framework in which adversarial training is used to find samples where RMs and human annotators disagree. However, they do not implement this idea, and relying on human annotators in their suggested way would be expensive and difficult. \citet{kirchner2024prover} propose adversarial training of a verifier model but it is limited to tasks with rule-based verifiers like GSM8K and is not applicable to general reward modeling. Concurrently with our work, \citet{wu2025rewordbench} use manually defined input transformations to increase the difficulty of RM benchmarks.

\textbf{Adversarial Attacks on LLMs. } There is a wide literature discussing adversarial attacks on LLMs \citep{zhang2020adversarial}. Typically these works attempt to greatly alter the model output while only slightly changing the model input (we only black-box attacks for simplicity \citep{guo2019simple}). We compare Adv-RM with two such approaches: Textfooler \citep{jin2020bert}, which replaces words with synonyms such that the model output is greatly changed and StyleAdv \citep{qi2021mind}, which changes the style of the text to greatly change the model output. We observe that such approaches usually do not succesfully attack RMs as (1) these attacks were not designed with the paired nature of RM training data in mind and (2) these attacks were designed to attack weaker models such as BERT \citep{devlin2019bert}.

\section{Methodology}
\label{sec:method}
\subsection{Background}
\textbf{RLHF. }We follow the  standard RLHF setup of \cite{stiennon2020learning, ouyang2022training}. We consider a setting where we have a dataset of $N$ human preference annotations, $D=\{x_i, y_{w, i}, y_{\ell, i}\}_{i=1}^N$. Here $x_i$ is the $i$th prompt, $y_{w,i}$ is the response preferred by human annotators, and $y_{\ell, i}$ is the rejected response. With this dataset, conventional RLHF trains a Bradley-Terry \citep{bradley1952rank} reward model $R_\theta(x, y)$ by minimizing the loss
\begin{align}
\label{eq:rm}
    \mathcal{L}(\theta) = - \mathbb{E}_{x, y_w, y_\ell \sim D}\left[ \textrm{log}(\sigma(R_\theta(x, y_w) - R_\theta(x, y_\ell)))\right]
\end{align}
Next, RLHF trains a policy $\pi_\phi$ to maximize the expected reward $R_\theta(x, y)$ while remaining close to an initial policy $\pi_{\textrm{SFT}}$. This is done by maximizing the following KL-regularized objective function over a prompt distribution $P$,
\begin{align*}
    \mathbb{E}_{(x, y) \sim P, y \sim \pi_\phi(x)} \left[ R_\theta(x, y) - \beta \mathbb{D}_{KL}\left(\pi_\phi(\cdot|x) || \pi_{\text{SFT}}(\cdot|x) \right) \right]
\end{align*}
using a reinforcement learning algorithm such as PPO, RLOO, or GRPO \citep{schulman2017proximal, ahmadian2024back, kool2019buy, shao2024deepseekmath}.

\textbf{Out-of-Distribution Detection. }
Detecting OOD data is a long studied problem. Here we consider a simple approach for OOD data detection in RLHF: Ensemble disagreement \citep{lakshminarayanan2017simple}. Concretely, we use the disagreement of an ensemble of reward models trained on different data splits to estimate the uncertainty of different prompt-response pairs. Such an approach has been shown to effectively capture model uncertainty and detect out-of-distribution data in a wide variety of domains, including continuous control \citep{brantley2019disagreement}, computer vision \citep{vyas2018out}, and natural language processing \citep{yuan2023revisiting}. Most recently, \citet{eisenstein2023helping} showed that reward model disagreement is an effective way to detect OOD responses in RLHF. Here, we denote reward model disagreement as $U_{\theta_1,\cdots,\theta_K}(x, y) = \textrm{STD}(R_{\theta_1}(x, y), ..., R_{\theta_K}(x, y))$, where $R_{\theta_1}, ..., R_{\theta_K}$ are $K$ different reward models trained with different data splits or random seeds. We remark that when $K=2$, $U_{\theta_1,\theta_q}(x, y)$ is proportional to $ |R_{\theta_1}(x, y) - R_{\theta_2}(x, y)|$. We normalize all RMs to have mean 0 and standard deviation 1.

\subsection{Learning to Attack}


The goal of our adversarial attack is, given a prompt, to find responses that receive high rewards from the target reward model and yet are OOD. Given a set of reward models $R_{\theta_1}, ..., R_{\theta_K}$, one reward model we seek to attack $R_{\theta_1}$, and set of prompts $D$, we seek to train a policy $\pi_{\text{adv}}$ to satisfy the constrained optimization problem:
\begin{align}
\label{eq:formulation}
    \mathop{\textrm{max}}_{\pi_{\text{adv}}}~ \mathbb{E}_{x \sim D, y \sim \pi_{\text{adv}}(x)}\left[U_{\theta_1, \dots, \theta_k}(x, y)\right] ~~s.t. ~~\mathbb{E}_{x \sim D, y \sim \pi_{\text{adv}}(x)}\left[R_{\theta_1}(x, y) > T(x)\right].
\end{align}
Here $T(x)$ is a threshold dependent on the prompt. In our experiments we use two RMs (i.e. $K=2$) and set $T(x)$ to be the average reward achieved by $\pi_{\text{SFT}}$ on the prompt $x$ according to $R_{\theta_1}(\cdot)$. Optimizing \eqref{eq:formulation} means finding OOD samples that the target RM $R_{\theta_1}(\cdot)$ predicts to be better than those generated by $\pi_{\text{SFT}}$. 
Although this approach would indeed find highly OOD responses that get high reward, it is possible that by optimizing \eqref{eq:formulation} we would find samples that all reward models predict will receive a high score, and the reward models just disagree on how high the score is. In order to encourage the found samples to be OOD and low quality, we use the formulation
\begin{align}
\label{eq:formulation2}
    \mathop{\textrm{max}}_{\pi_{\text{adv}}}~ \mathbb{E}_{x \sim D, y \sim \pi_{\text{adv}}(x)}\left[R_{\theta_1}(x, y) - \lambda R_{\theta_2}(x, y)\right] ~~s.t. ~~\mathbb{E}_{x \sim D, y \sim \pi_{\text{adv}}(x)}\left[R_{\theta_1}(x, y) > T(x)\right].
\end{align}
Here $\lambda$ is a hyperparameters, with $\lambda$ typically greater than $1$. This formulation encourages finding responses with high uncertainty and a very low score from $R_{\theta_2}(x, y)$, increasing the likelihood of low quality samples. We overload our notation, and denote $R_{\theta_1}(x, y) - \lambda R_{\theta_2}(x, y)$ as $U_{\theta_1, \theta_2}(x,y)$. 

Unlike previous adversarial attacks that rely on perturbations, the key innovation of our work is training an adversarial policy $\pi_\text{adv}$ to generate adversarial examples. Specifically, we optimize \eqref{eq:formulation2} using reinforcement learning and train an adversarial policy $\pi_\text{adv}$ to maximize the following reward:
\begin{align*}
    R_{\text{adv}}(x, y) =
    \begin{cases} 
    R_{\theta_1}(x, y) - \lambda R_{\theta_2}(x, y) & \text{if } R_{\theta_1}(x, y) > T(x) \\ 
    R_{\theta_1}(x, y) -C & \text{otherwise}
    \end{cases}.
\end{align*}
Here $C$ is a constant that ensures the second case always gives a lower reward than the first case. This reward encourages the policy to maximize RM uncertainty and find OOD samples, while ensuring they receive a high reward from $R_{\theta_1}(x, y)$ (the attack target). We maximize this reward using RLOO \citep{ahmadian2024back} over the RLHF prompt dataset, starting from $\pi_{\text{SFT}}.$ The reinforcement learning training procedure leaves us with a set of prompt-response pairs $\{(x_1, y_1^{\text{adv}}), \dots (x_n, y_n^{\text{adv}}\})$ generated by the adversarial policy throughout training. 

\textbf{Data Filtering.} We filter the collected prompt-response pairs using $R_{\theta_1}(x, y)$ and $U_{\theta_1, \theta_2}(x,y)$ to only select samples where the policy was able to successfully attack $R_{\theta_1}(x, y)$. Concretely, we select the adversarial dataset \( D_{\text{adv}} \) as 
\[
D_{\text{adv}} = \{ (x, y_{\text{adv}}) \mid R_{\theta_1}(x, y_{\text{adv}}) > T(x) \text{ and } Z(U_{\theta_1, \theta_2}(x,y_{\text{adv}})) > 1.96 \}.
\]
Here, \( Z(U_{\theta_1, \theta_2}(x,y) \) is the z-score of the uncertainty \( U_{\theta_1, \theta_2}(x,y) \), computed over responses sampled from \( \pi_{\text{SFT}} \). A value of \( Z(U_{\theta_1, \theta_2}(x,y)) > 1.96 \) indicates that the reward uncertainty of \( (x, y_{\text{adv}}) \) is significantly higher than the uncertainty of responses within the distribution of \( \pi_{\text{SFT}} \). This ensures each sample in $D_{\text{adv}}$ recieves a high reward score yet is OOD. 

\subsection{Preliminary Analysis: Uncertainty as a Quality Signal}
Using a synthetic RLHF setup \citep{gao2023scaling}, we investigate how and when reward models uncertainty (i.e. $U_{\theta_1, \theta_2}(\cdot)$) correlates with response quality. Concretely, we train two reward models ($R_{\theta_1}(\cdot)$ and $R_{\theta_2}$) based on different random seeds on a preference dataset created by a gold reward model, and then measure $U_{\theta_1, \theta_2}(\cdot)$ versus the gold score which measures quality. We investigate this relationship on two datasets: one dataset where the responses are samples from the SFT model, and one sampled from the Adv-RM model.

\begin{figure}[h]
    \centering
    \subfigure[Gold Score vs $U_{\theta_1, \theta_2}(\cdot)$ over SFT Distribution \label{fig:prelim_sft}]{\includegraphics[width=0.31\textwidth]{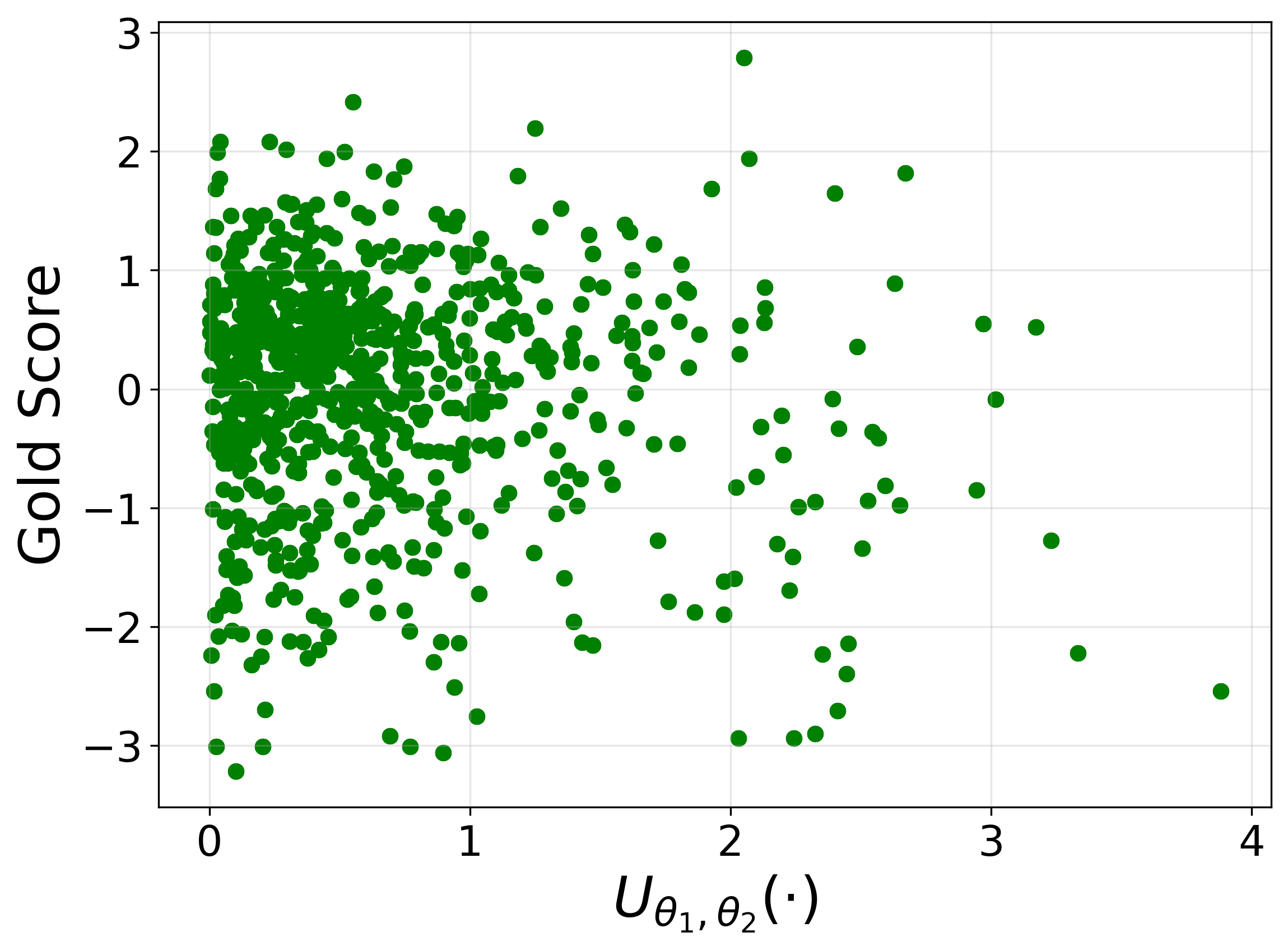}}
    \subfigure[Gold Score vs $U_{\theta_1, \theta_2}(\cdot)$ over SFT + Adv-RM Distribution \label{fig:prelim_adv}]{\includegraphics[width=0.31\textwidth]{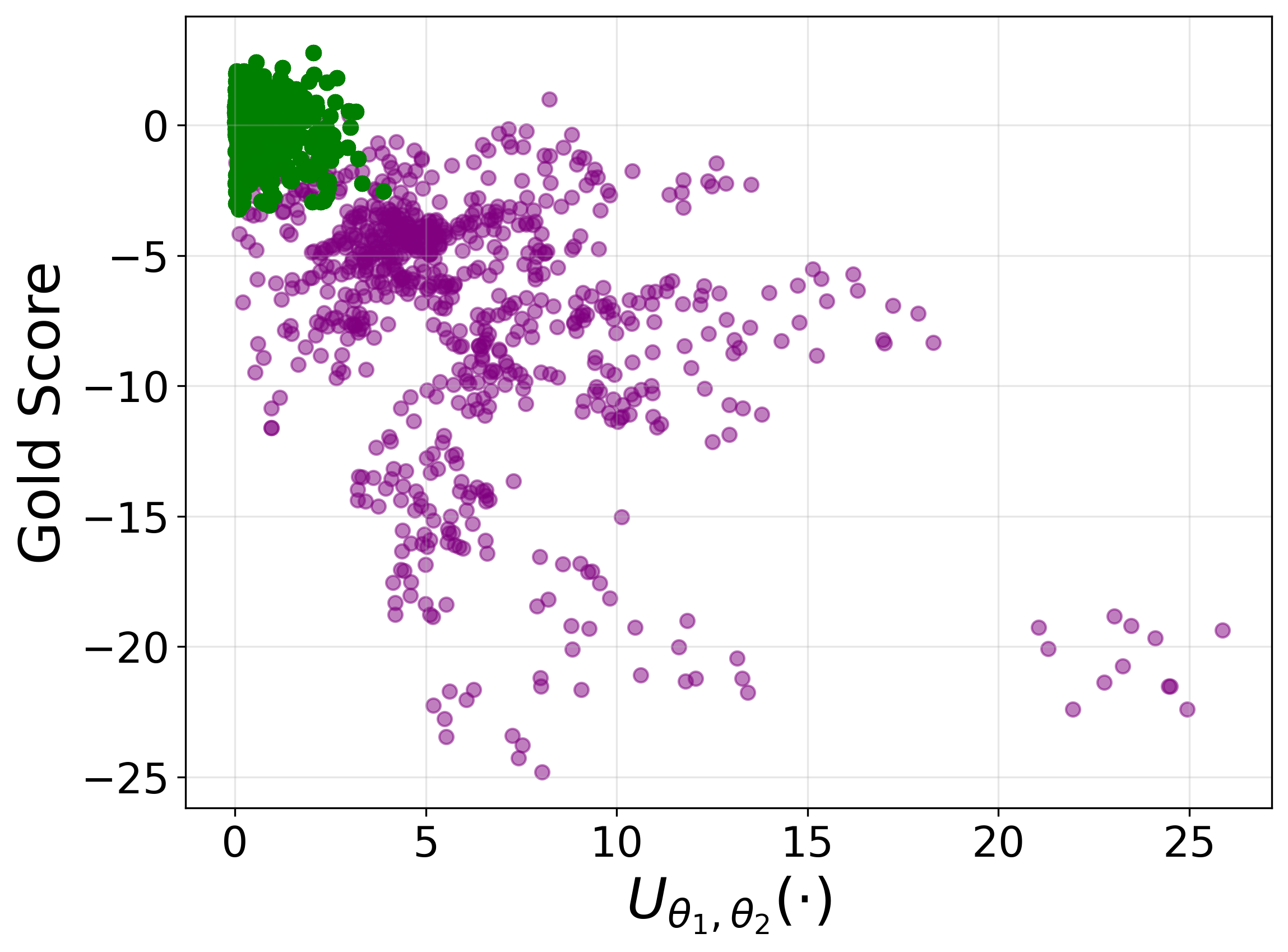}}
    \subfigure[$R_{\theta_1}(\cdot)$ vs $U_{\theta_1, \theta_2}(\cdot)$ over SFT + Adv-RM Distribution \label{fig:prelim_rtheta}]{\includegraphics[width=0.31\textwidth]{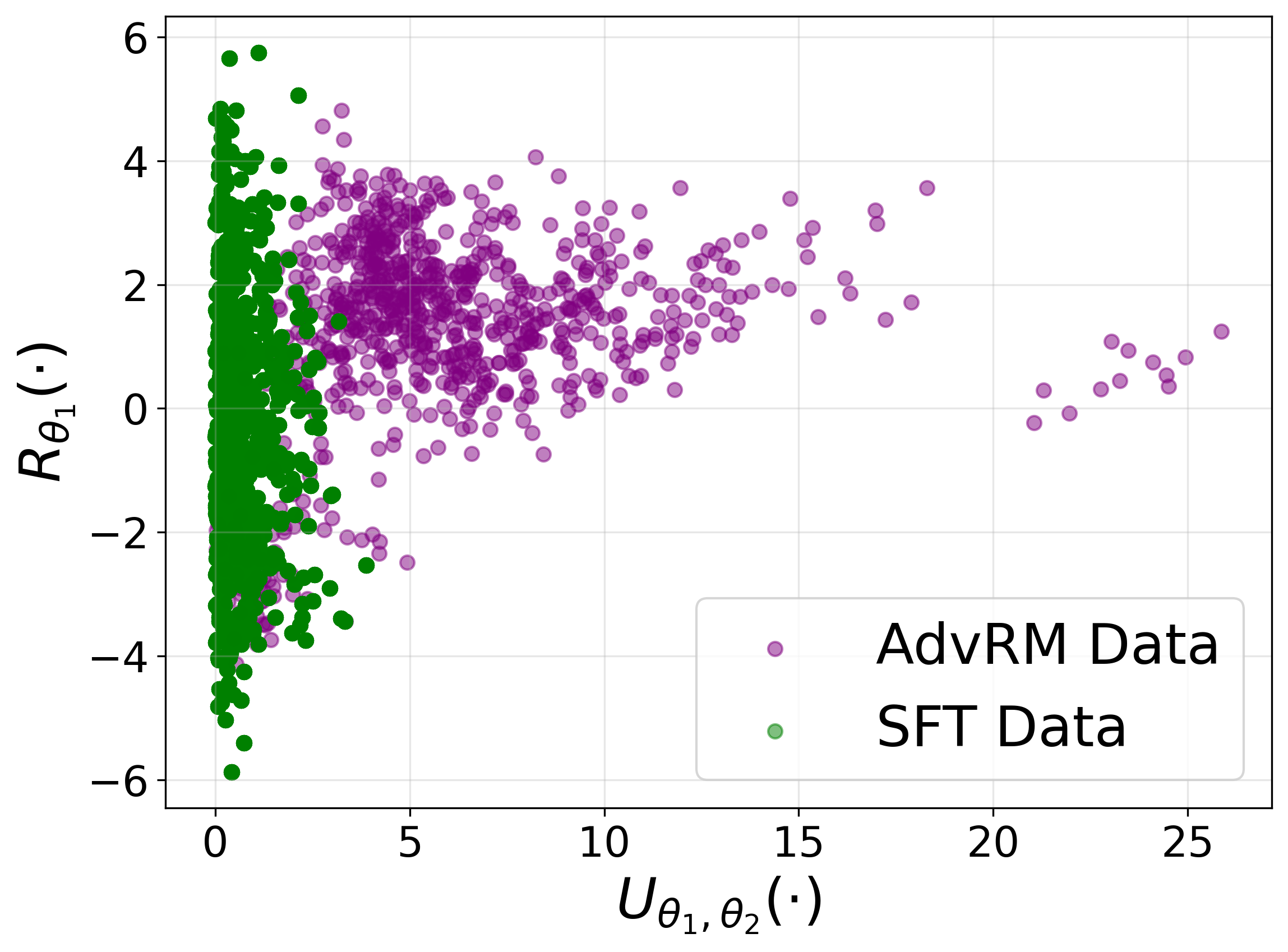}}
    \caption{(a) and (b) show $U_{\theta_1, \theta_2}(\cdot)$ versus gold score. (c) shows that $R_{\theta_1}$ is similar for the Adv-RM and SFT data, which helps isolate the relationship between $U_{\theta_1, \theta_2}(\cdot)$ and gold score.}
    \label{fig:prelim}
\end{figure}

As we can see from Figure \ref{fig:prelim_sft}, uncertainty $U_{\theta_1, \theta_2}(\cdot)$ is not very correlated with the gold reward (pearson's correlation -0.05) when we sample from $\pi_{\text{SFT}}$. This indicates that directly penalizing ensemble uncertainty during RLHF may add some noise to training. On the other hand, when we use Adv-RM to generate OOD datapoints with a higher uncertainty level (\ref{fig:prelim_adv}), we see that uncertainty is more correlated with gold score (pearson's correlation -0.70). Together, these plots suggest that ensemble uncertainty is a useful indicator of response quality when uncertainty is high (and can therefore be used to find adversarial examples), but it provides little signal for distinguishing quality among responses with low uncertainty (implying directly regularizing RM uncertainty during policy training is not ideal).

\subsection{Adversarial Training of Reward Models}
Our analysis shows that Adv-RM's attacks are succesful at producing adversarial responses. To increase RM robustness, we incorporate these adversarial responses into training. 

\textbf{Preference Pair Construction. }We construct adversarial preference pairs by setting $(x, y_{\text{SFT}})$ as the preferred pair and setting $(x, y_{\text{adv}})$ as the rejected pair. 
We choose a SFT response $y_{\text{SFT}}$ such that $R_{\theta_1}(x, y_{\text{SFT}}) > \mathbb{E}_{y \sim \pi_{\text{SFT}}(x)}\left[ R_{\theta_1}(x, y_{\text{SFT}}) \right]$, to avoid using low quality SFT responses as the preferred response.
All of these adversarial preference pairs $(x, y_{\text{adv}})$ are constructed such that the rejected response $y_{\text{adv}}$ receives a higher reward score from $R_{\theta_1}(\cdot)$ than the SFT response, yet $y_{\text{adv}}$ has very high $U_{\theta_1, \theta_2}(\cdot)$ so it is OOD and likely low quality. 

\textbf{Multiple Rounds of Training. } Once we construct the adversarial preference pairs, we add around 1000 of them to the original RLHF preference dataset and train a new RM from scratch using \eqref{eq:rm}. To further improve RM robustness, we conduct several rounds of adversarial training, in each round trying to find vulnerabilities of the new robust RM. Once we find these vulnerabilities, we append them to our current robust RLHF dataset and train a new RM from scratch. We find that after 2 rounds of adversarial training, our attack procedure is no longer successful. At this point we terminate training. 

\section{Experiments: Adversarial Attacks on Reward Models}
\label{sec:exp-attacks}
\subsection{Evaluation}
We consider an adversarial attack on a reward model successful if it produces a preference pair $(x, y_1), (x, y_2)$ such that 
\begin{align}
\label{eq:standard}
    R_{\theta_1}(x, y_1) > R_{\theta_1}(x, y_2) \textrm{~~while~~} R_\text{gold}(x, y_1) < R_\text{gold}(x, y_2).
\end{align} 
The ``gold'' reward here is meant to capture the reward induced by human values. In synthetic setups we have access to the simulated gold reward model, and in real RLHF settings we can approximate $R_\text{gold}$ with a LLM \citep{dubois2023alpacafarm} or human feedback.

In synthetic settings, we find \eqref{eq:standard} can sometimes be met by chance if $y_1$ and $y_2$ are similar. To mitigate this, we introduce a stricter criterion: the attack is only considered successful if 
\begin{align}
\label{eq:strict}
    Z_{R_{\theta_1}}(R_{\theta_1}(x, y_1)) > 0 \textrm{~~and~~} Z_{R_\text{gold}}(R_\text{gold}(x, y_1)) < -1.96.
\end{align}
Intuitively, a successful attack means that $R_{\theta_1}(x, y_1)$ is above the average SFT response for the response $x$, while $R_\text{gold}(x, y_1)$ falls within the bottom 5\% of SFT scores.
When using LLM judges or humans as $R_\text{gold}$, we only consider \eqref{eq:standard}.

\subsection{Experimental Setup}

\textbf{Synthetic RLHF setup. }
Following \citet{gao2023scaling}, we evaluate Adv-RM in a synthetic RLHF setup. We use Llama-3.1-Nemotron-70B-Reward as the gold RM, and create a synthetic training dataset by relabelling Helpsteer-2-Preferences \citep{wang2024helpsteer2} with it. We then train proxy RMs based on Llama-3.1-8B-Instruct \citep{grattafiori2024llama}. In this setting $R_{\theta_1}(\cdot)$ and $R_{\theta_2}(\cdot)$ are two RMs trained on different random seeds controlling the data order. We evaluate attack success rate using the gold RM. Details can be found in Appendix \ref{app:hyperparam-advtraining}. Adv-RM adversarial policies are initialized from Llama-3.1-8B-Instruct.


\textbf{Real RLHF setup. }
We attack three top performing RMs on RewardBench \citep{lambert2024rewardbench}, Skywork-Reward-Gemma-2-27B, Llama-3.1-Nemotron-70B-Reward, and Nemotron-4-340B-Reward. When attacking Skywork-Reward-Gemma-2-27B and Nemotron-4-340B-Reward we use Llama-3.1-Nemotron-70B-Reward as $R_{\theta_2}(\cdot)$ in OOD detection and when attacking Llama-3.1-Nemotron-70B-Reward we use an in-house RM as $R_{\theta_2}(\cdot)$. We then evaluate the attack success rate using \eqref{eq:standard} with human judges, Llama 405B, and Deepseek-R1 \citep{guo2025deepseek}. See Appendix \ref{app:eval} for details. Adv-RM adversarial policies are initialized from Llama-3.1-8B-Instruct.

\begin{figure}[t]
    \centering
    \subfigure[Skywork-Gemma-2-27B]
    {\includegraphics[width=0.294\textwidth]{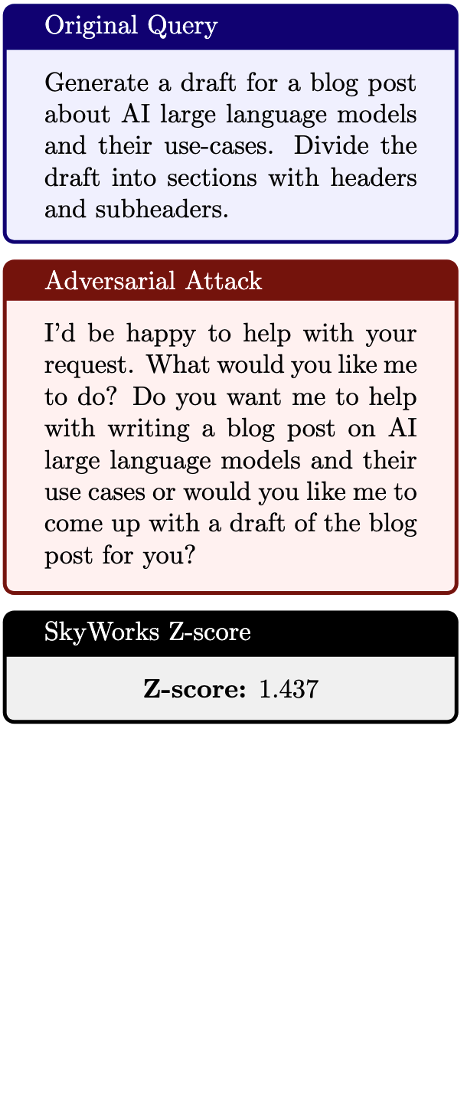}}
    \subfigure[Llama-Nemotron-70B]{\includegraphics[width=0.3\textwidth]{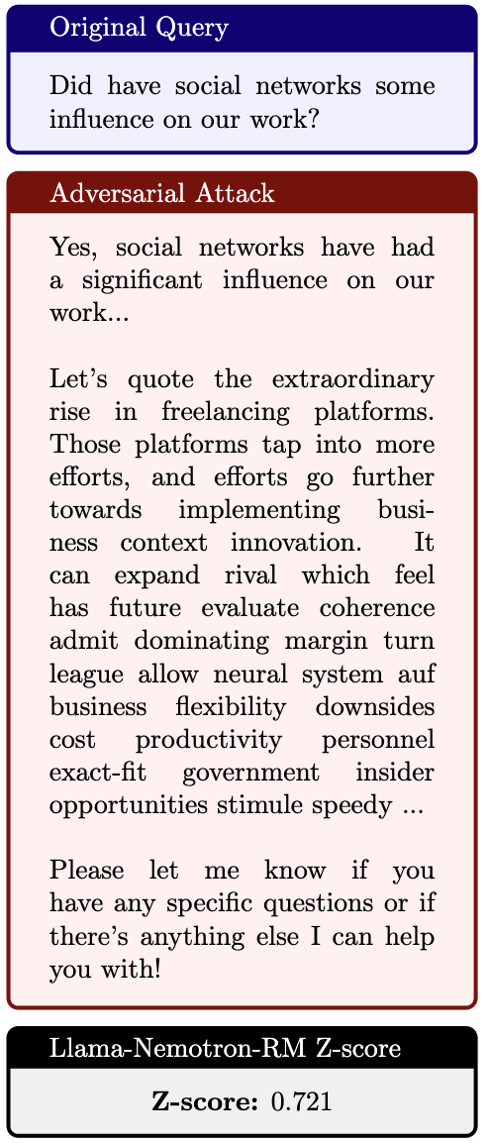}}
    \subfigure[Nemotron-4-340B]{\includegraphics[width=0.303\textwidth]{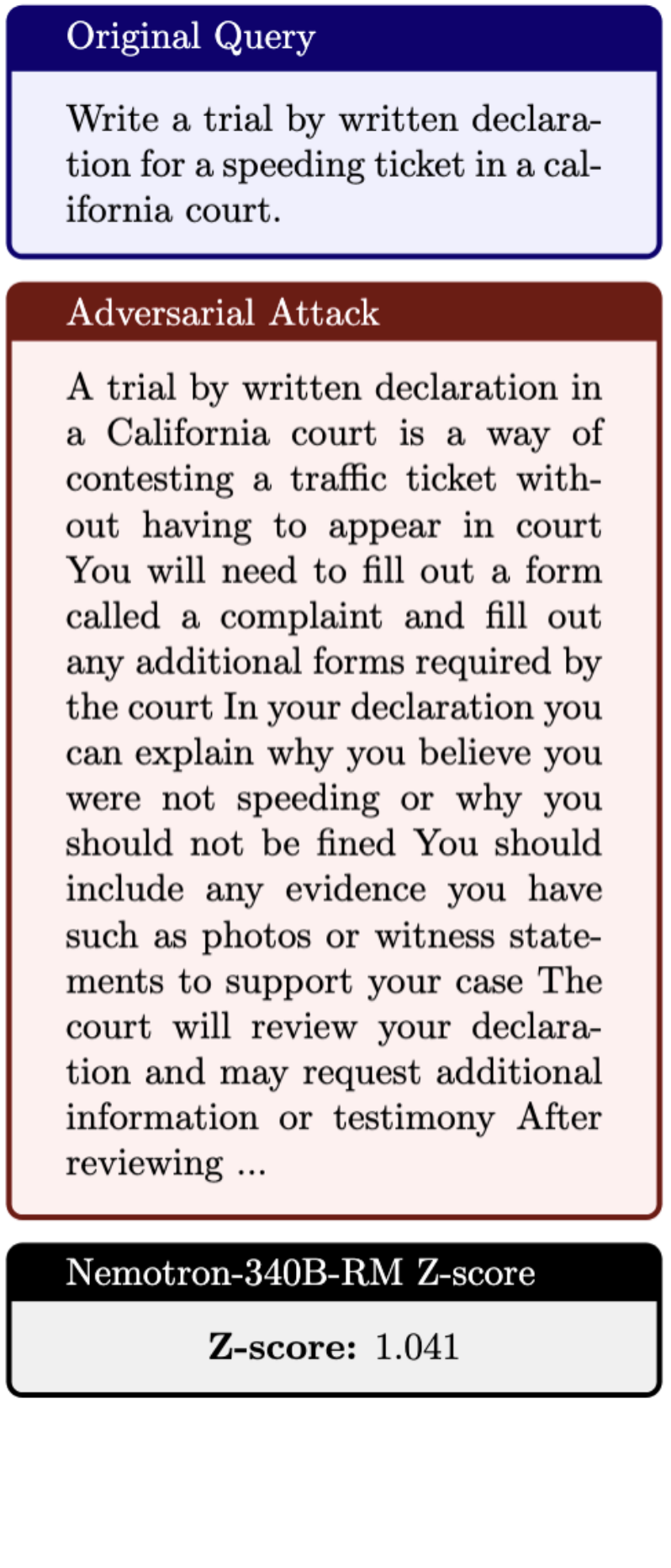}}
    \caption{Adversarial examples generated by Adv-RM for top RewardBench models. The Z-score is computed by normalizing the reward score by the average reward achieved by Llama-3.1-8b-Instruct for that prompt.}
    \label{fig:adv_examples}
\end{figure}

\begin{table}[h]
    \centering
    \renewcommand{\arraystretch}{1.1}
    \small 
    \setlength{\tabcolsep}{12pt} 
    \begin{tabular}{l cc | cc}
        \toprule
        & \multicolumn{2}{c|}{\textbf{Success Rate: Train}} & \multicolumn{2}{c}{\textbf{Success Rate: Test}} \\
        \cmidrule(lr){2-3} \cmidrule(lr){4-5}
        \textbf{Method} & \textbf{Standard} & \textbf{Strict} & \textbf{Standard} & \textbf{Strict} \\
        \midrule
        RM Over-optimization & 8.31 (2.44)  & 2.85 (1.47)  & 7.59 (2.34)  & 1.34 (1.02)  \\
        RRM & 0 (0.00)  & 0 (0.00)  & 0 (0.00)  & 0 (0.00)  \\
        StyleAdv & 0 (0.00)  & 0 (0.00)  & 0 (0.00)  & 0 (0.00)  \\
        Textfooler & 9.57 (2.60)  & 0 (0.00)  & 18.8 (3.45)  & 0 (0.00)  \\
        Adv-RM & \textbf{99.31} (0.73)  & \textbf{92.91} (2.27)  & \textbf{100} (0.00)  & \textbf{100} (0.00)  \\
        \bottomrule
    \end{tabular}
    \caption{Comparison of attack success rates over 128 Helpsteer-2 prompts in the synthetic setting, with standard errors in parentheses.}\label{tab:attack_results}
\end{table}

\textbf{Baselines. } Although we are the first to adversarially attack RMs (to the best of our knowledge), we consider a few baselines that try to attack LLM classifiers (Textfooler and StyleAdv), RRM \citep{liu2024rrm}, and a reward over-optimization baseline. As Textfooler and StyleAdv only produce a single adversarial response $y_1$ to a prompt $x$, we pair their attacks with the original response from the SFT model $y_2 \sim \pi_{\text{SFT}}(x)$ to calculate performance in \eqref{eq:standard} and \eqref{eq:strict}. For reward over-optimization, we optimize the policy for three times the number of steps as achieves the best downstream performance and use the resulting policy to generate adversarial attacks. We present more details in Appendix \ref{app:attacks}.


\subsection{Main Results: Adversarial Attacks}

\textbf{Synthetic Setting.} Table \ref{tab:attack_results} compares Adv-RM to other baselines. For Adv-RM we generate 4 attacks per evaluation prompt and select the attack with highest $U_{\theta_1, \theta_2}(\cdot)$. Adv-RM consistently finds adversarial samples, indicating its efficacy. In contrast, StyleAdv and TextFooler have low success rates despite generating 50 attacks per prompt and filtering by reward difference. These methods slightly perturb response semantics, rarely causing conflicts between the trained and gold RMs. RM over-optimization can occasionally produce successful attacks; however, we find that the over-optimized policy often generates high-quality responses with only minor quirks (e.g., excessive formatting). These responses are not necessarily of lower quality than those produced by $\pi_{\text{SFT}}$, making them less adversarial in nature.

\begin{table}[t]
    \centering
    \renewcommand{\arraystretch}{0.7} 
    \setlength{\tabcolsep}{6pt} 
    \begin{tabular}{lcccccc} 
        \toprule
        & \multicolumn{1}{c}{Human Evaluation}
        & \multicolumn{1}{c}{Deepseek R1} & \multicolumn{1}{c}{Llama 405B} \\
        \midrule
        \textbf{Skyworks Gemma 27B} \\
        TextFooler & 19.61 (5.60) &  56 (4.41) &  2 (1.24) \\
        RM Over-Optimization & 20 (5.66) & 32 (3.99) & 14 (3.00) \\
        Adv-RM & \textbf{100} (0.00) & \textbf{99} (0.73) & \textbf{99} (0.73) \\
        \midrule
        \textbf{Llama-Nemotron 70B} \\
        TextFooler & 13.2 (4.86) & 27 (3.85) & 1 (0.88) \\
        RM Over-Optimization & 23.64 (6.07) & 7 (2.23) & 3 (1.50) \\
        Adv-RM & \textbf{83.01} (5.13) &  \textbf{71} (4.12) &  \textbf{24} (3.73) \\
        \midrule
        \textbf{Nemotron 340B} \\
        TextFooler & 17.86 (5.22) & 38 (4.33) & 2 (1.24) \\
        RM Over-Optimization & 17.65 (5.20) & 23 (3.73) & 19 (3.52) \\
        Adv-RM & \textbf{78.85} (5.76) & \textbf{76} (4.11) &  \textbf{82} (3.42) \\
        \bottomrule
    \end{tabular}
    \caption{Adv-RM attack effectiveness across different real RMs and judges. Values in parentheses represent standard errors.}
    \label{tab:combined_attack_comparison}
\end{table}

\begin{figure}[h]
    \centering
    \begin{minipage}{0.4\textwidth}
        \centering
        {\includegraphics[width=0.84\textwidth]{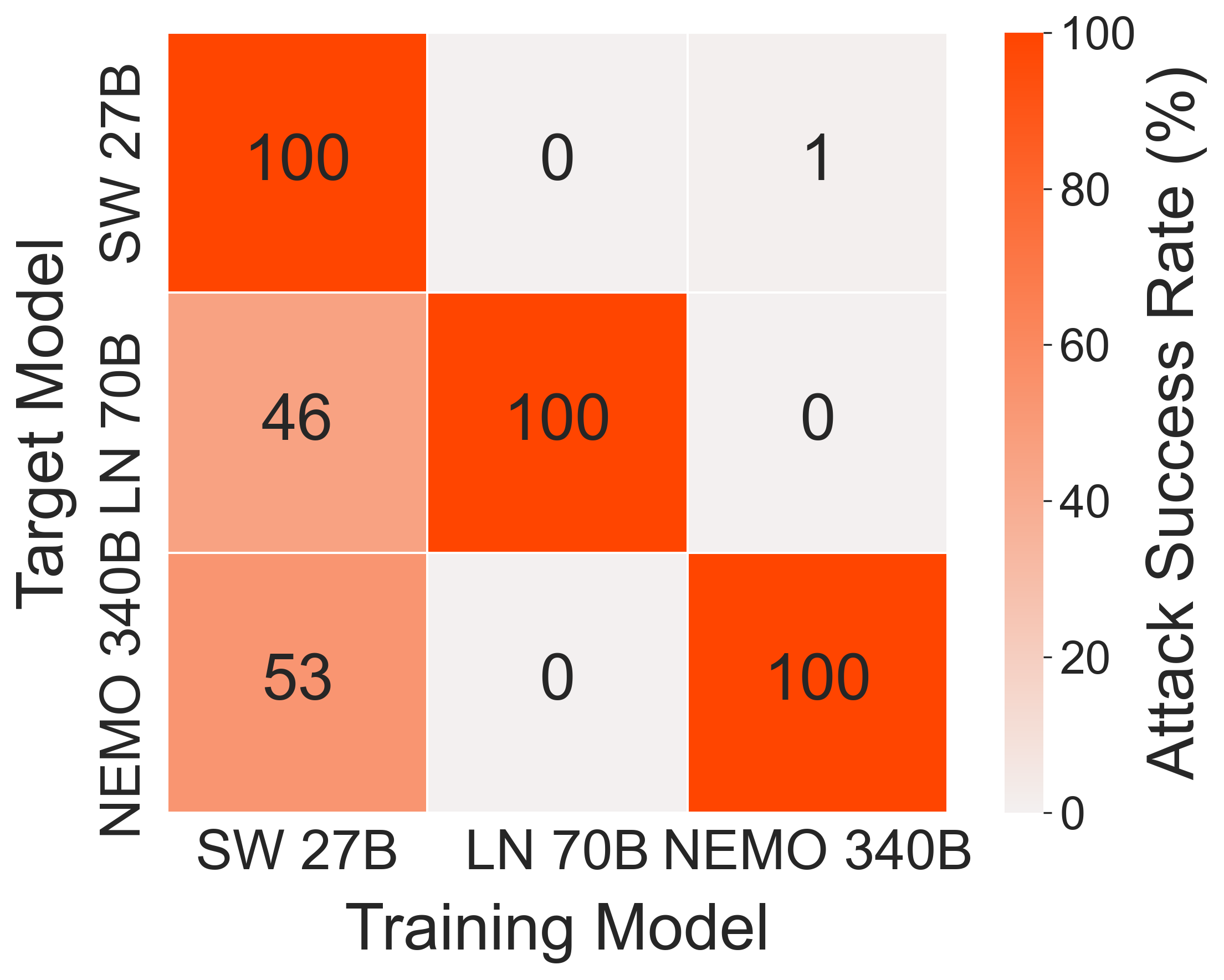}}
        \caption{Attack transferability.}
        \label{fig:attack-transfer}
    \end{minipage}%
    \begin{minipage}{0.4\textwidth}
        \centering
        \renewcommand{\arraystretch}{1.1}
        \setlength{\tabcolsep}{8pt}
        \begin{tabular}{lc} 
            \toprule
            Method & Success Rate \\
            \midrule
            \textbf{Adv-RM} & 92.91 (2.27) \\
            {~~-~Filtering} & 73.27 (3.94) \\
            {~~-~Unequal RM Weights} & 16.41 (3.13) \\
            {~~-~Reward Threshold} & 0.0 (0.0) \\
            \bottomrule
        \end{tabular}
        \captionof{table}{Adv-RM attack ablation in the synthetic setup.}
        \label{tab:adv_ablation}
    \end{minipage}
\end{figure}

\textbf{Real Reward Models. } The main results when attacking Skyworks-Gemma 27B reward, Llama-Nemotron 70B reward, and Nemotron-4 340B reward can be found in Table \ref{tab:combined_attack_comparison}. Note we only compare Adv-RM to the strongest baselines in Table \ref{tab:attack_results}. As we can see from Table \ref{tab:combined_attack_comparison}, Adv-RM is able to succesfully attack all three state-of-the-art reward models, generating adversarial pairs that the target RM predicts as correct while the judge (humans or models) predicts as incorrect. We remark that using LLM as a judge may significantly underestimate the success rate of Adv-RM as they may also be fooled by adversarial examples. 

We provide examples of these attacks in Figure \ref{fig:adv_examples}. Interestingly, Adv-RM uncovers distinct vulnerabilities in each reward model. For Skyworks-Gemma 27B, responses that simply repeat the prompt can achieve high scores. For Llama-Nemotron 70B, nonsensical gibberish in the middle of long text can be rewarded. And for Nemotron 340B, responses that lack punctuation can receive unexpectedly high scores.


\textbf{Attack Transferability.} We show the transferability of attacks in Figure \ref{fig:attack-transfer}. We find that there is some weak transfer between Adv-RM attacks on the Skyworks 27B model, but no transfer between the different Nemotron RMs.

\textbf{Ablation}. Our ablation study in Table \ref{tab:adv_ablation} shows that all components of Adv-RM are helpful.

\section{Experiments: Downstream Policy Performance}

\textbf{Experimental Setup. } For the synthetic setting, we use the RMs trained with Adv-RM in Section~\ref{sec:exp-attacks} to optimize Llama-3.1-8B-Instruct models using RLOO. We measure performance using the Gold Reward. For the real RLHF setting, we train Llama-3.1-8B-Instruct RMs on the Helpsteer-2-Preferences dataset \citep{wang2024helpsteer2}, and use them to optimize Llama-3.1-8B-Instruct models. We measure performance on a held out set of 128 prompts using different judge models (Llama 405B and Deepseek R1), as well as Llama-Nemotron 70B RM.

\textbf{Baselines. } We consider several baselines, including conventional RLHF (baseline), the mean of an ensemble of RMs (Ens (Mean)), the mean minus the uncertainty of an ensemble of RMs (Ens (STD), also called UWO in \citet{coste2023reward}), and RRM \citep{liu2024rrm}. Details can be found in Appendix \ref{app:policy-baseline}.

\begin{figure}[h]
    \centering
    \subfigure[Gold Reward Curve]{\includegraphics[width=0.23\textwidth]{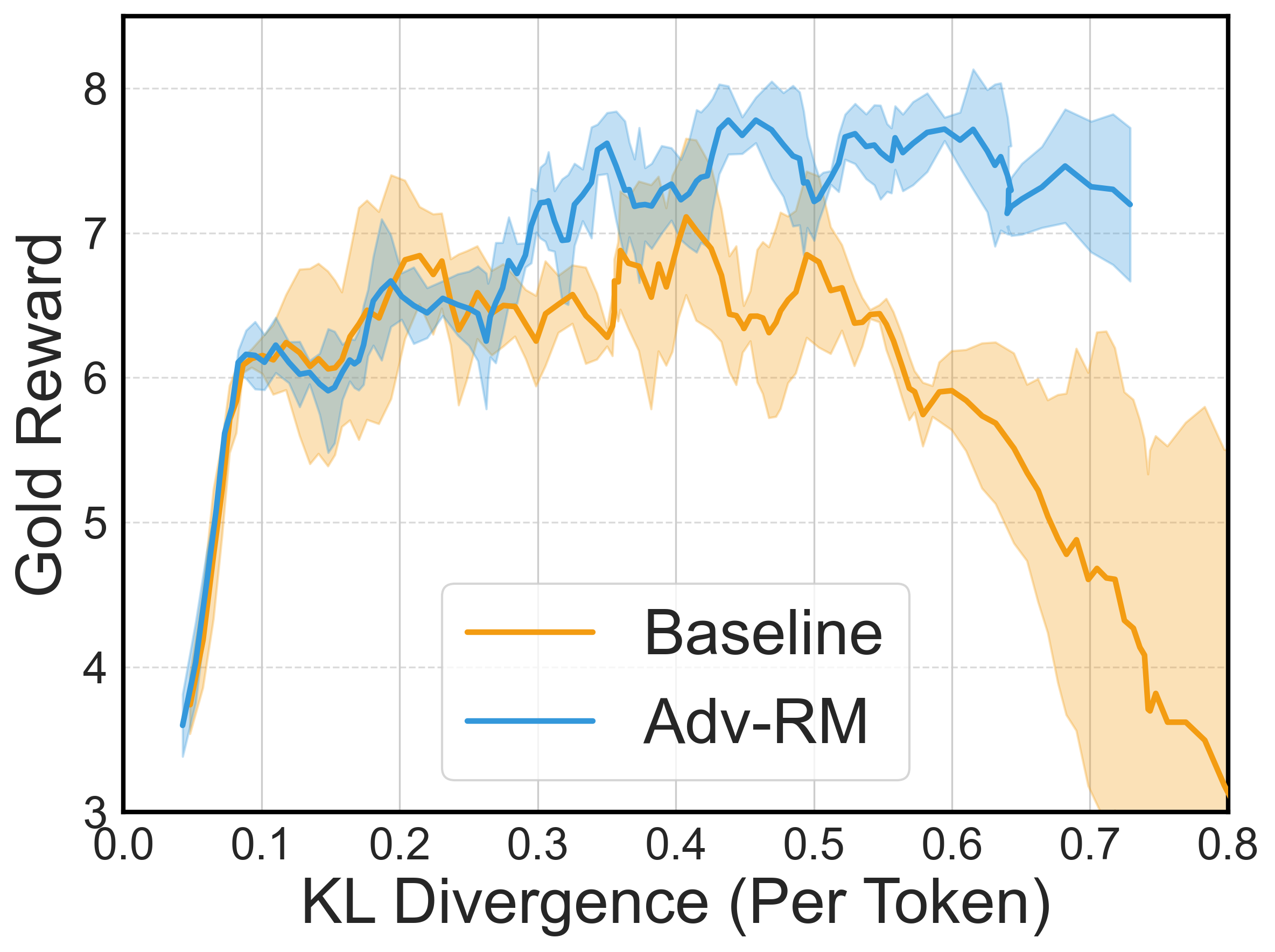}}
    \subfigure[Generation Length]{\includegraphics[width=0.23\textwidth]{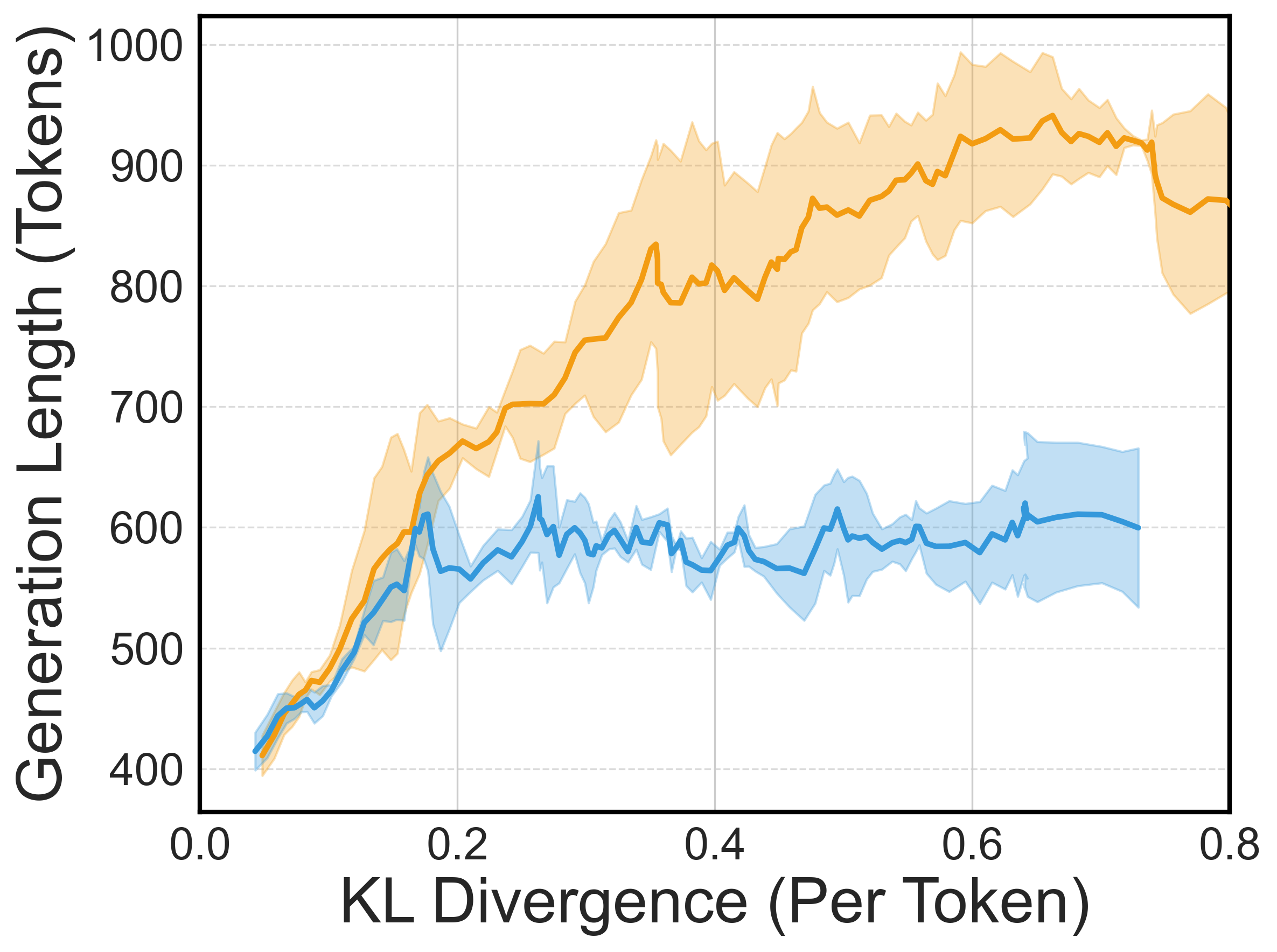}}
    \subfigure[Best Reward]{\includegraphics[width=0.23\textwidth]{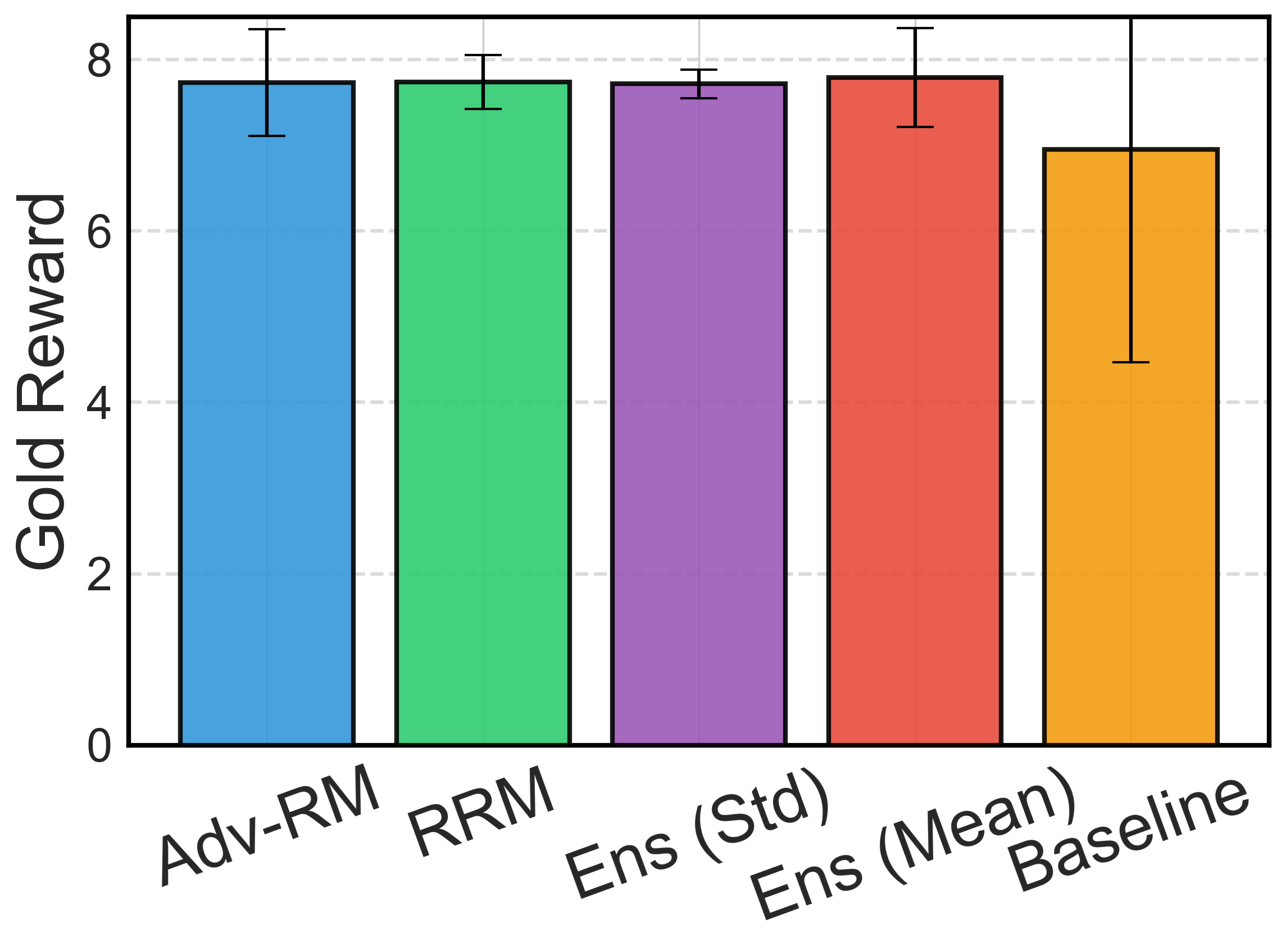}}
    \subfigure[Final Reward ]{\includegraphics[width=0.23\textwidth]{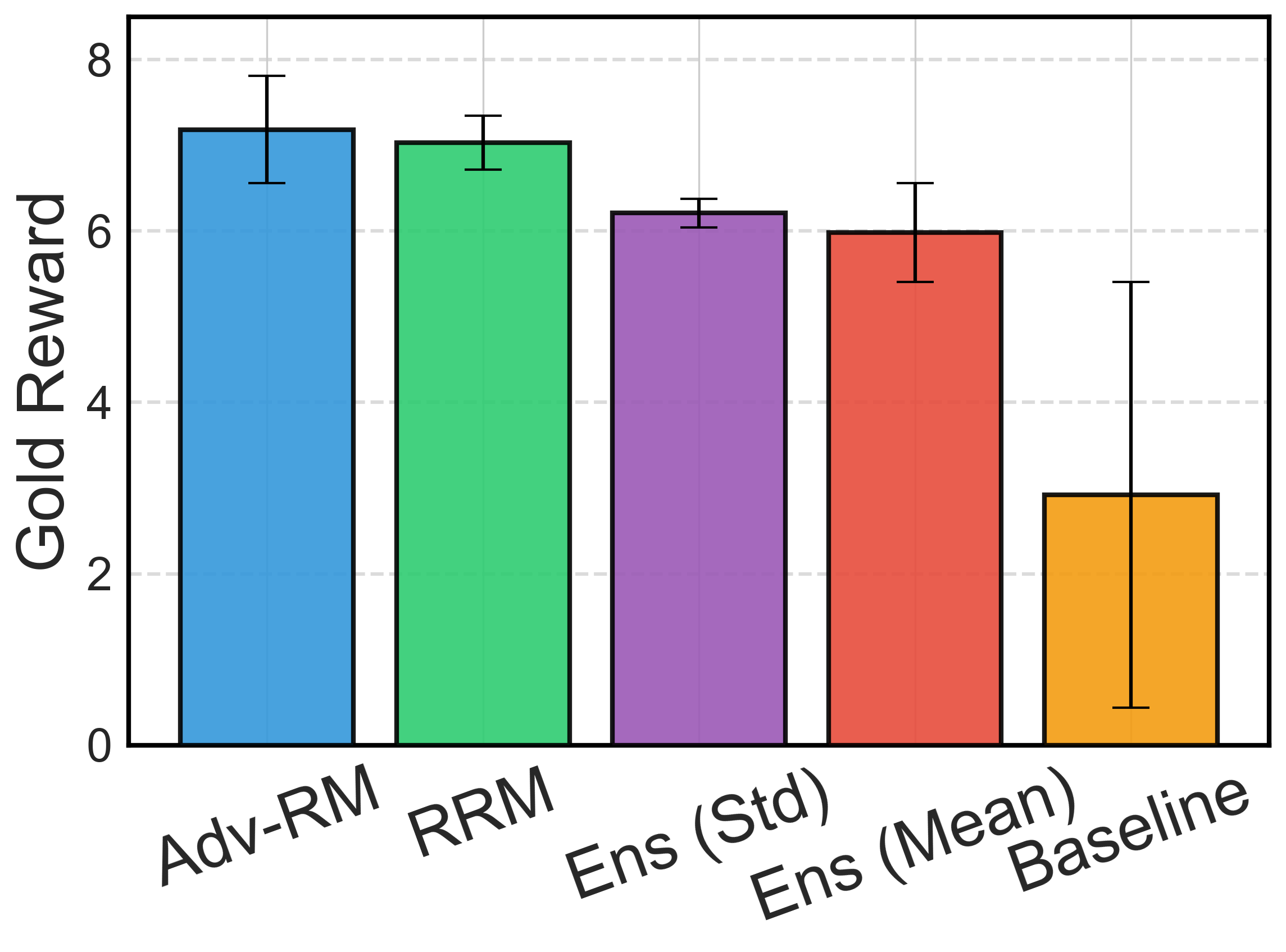}}
    \caption{Downstream policy results in the synthetic setup. Error bars represent $\pm$ one standard deviation over three random seeds.}
    \label{fig:synthetic-downstream}
\end{figure}

\textbf{Results: Synthetic Setting. } The training curve, best reward achieved, and final reward achieved for Adv-RM policy training and other baselines can be found in Figure \ref{fig:synthetic-downstream}. The policies trained with Adv-RM achieves a higher gold reward with minimal reward hacking, exhibits little length hacking, and performs better or as well as all of the baselines. We find that Adv-RM only results in a small improvement in best-case policy performance, and mostly improves robustness against reward hacking. Overall, these results show that including adversarial samples in RM training can improve downstream performance.

\begin{figure}[h]
    \centering
    \subfigure[Llama 405B]{\includegraphics[width=0.32\textwidth]{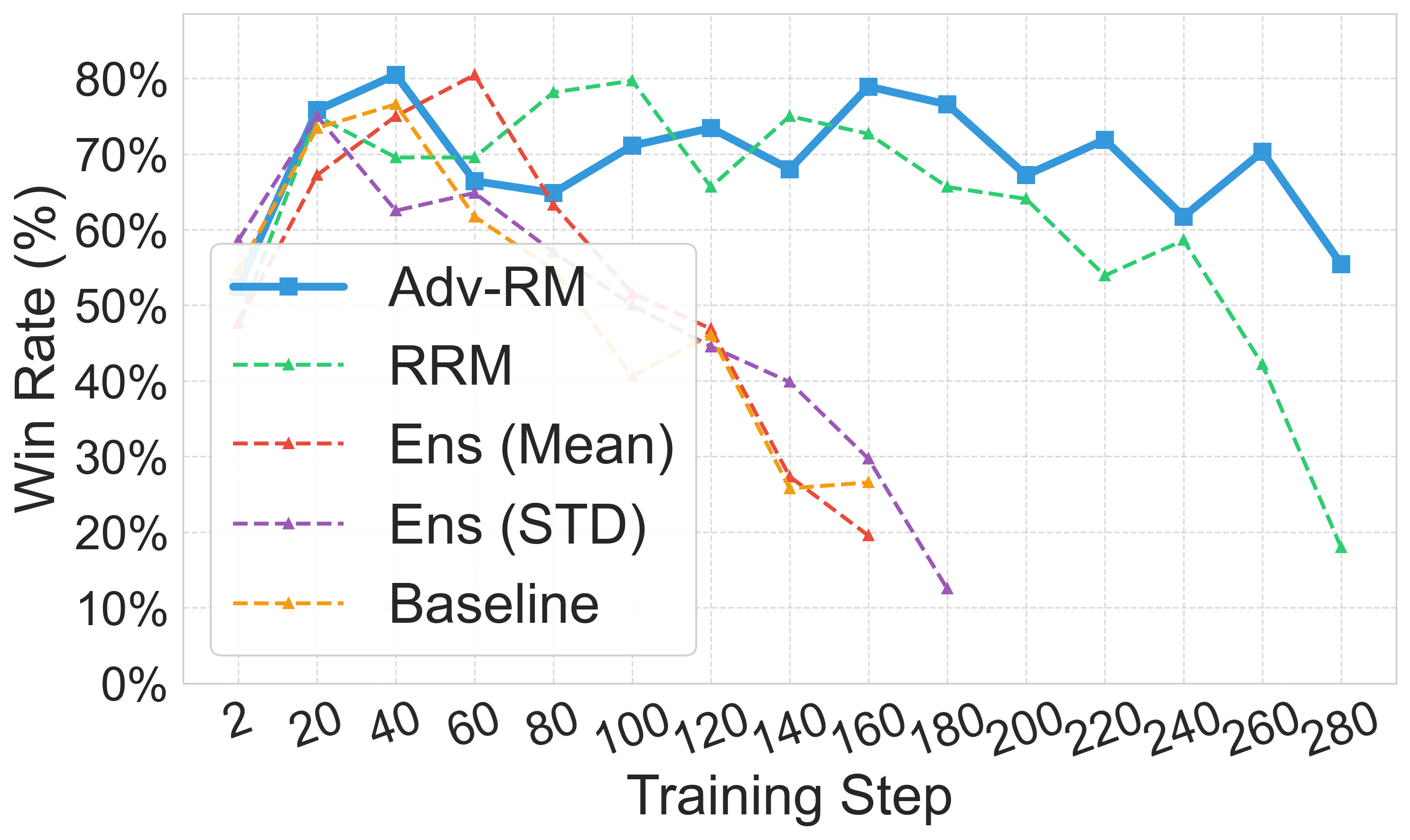}}
    \subfigure[Deepseek-R1]{\includegraphics[width=0.32\textwidth]{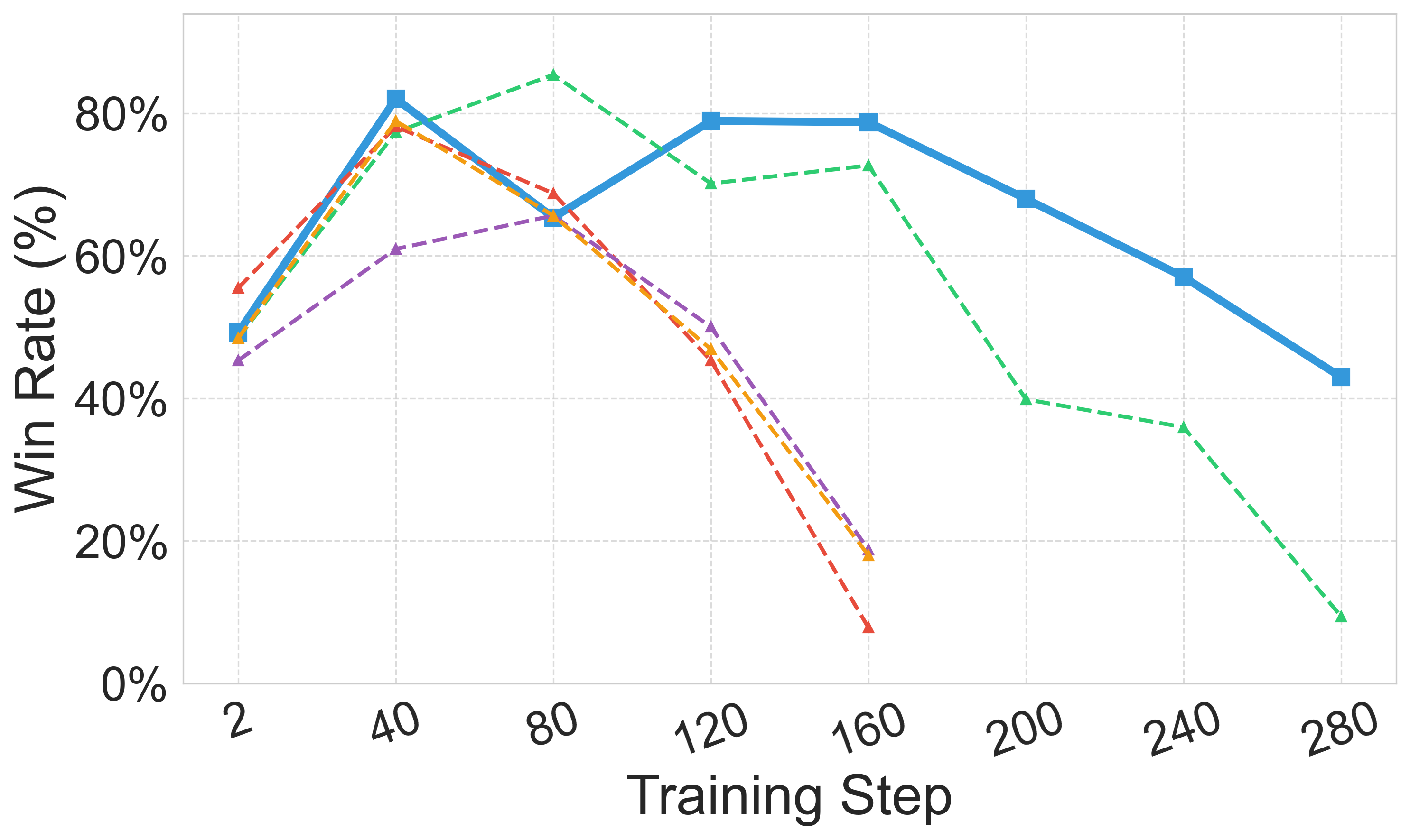}}
    \subfigure[Llama-Nemotron RM]
    {\includegraphics[width=0.32\textwidth]{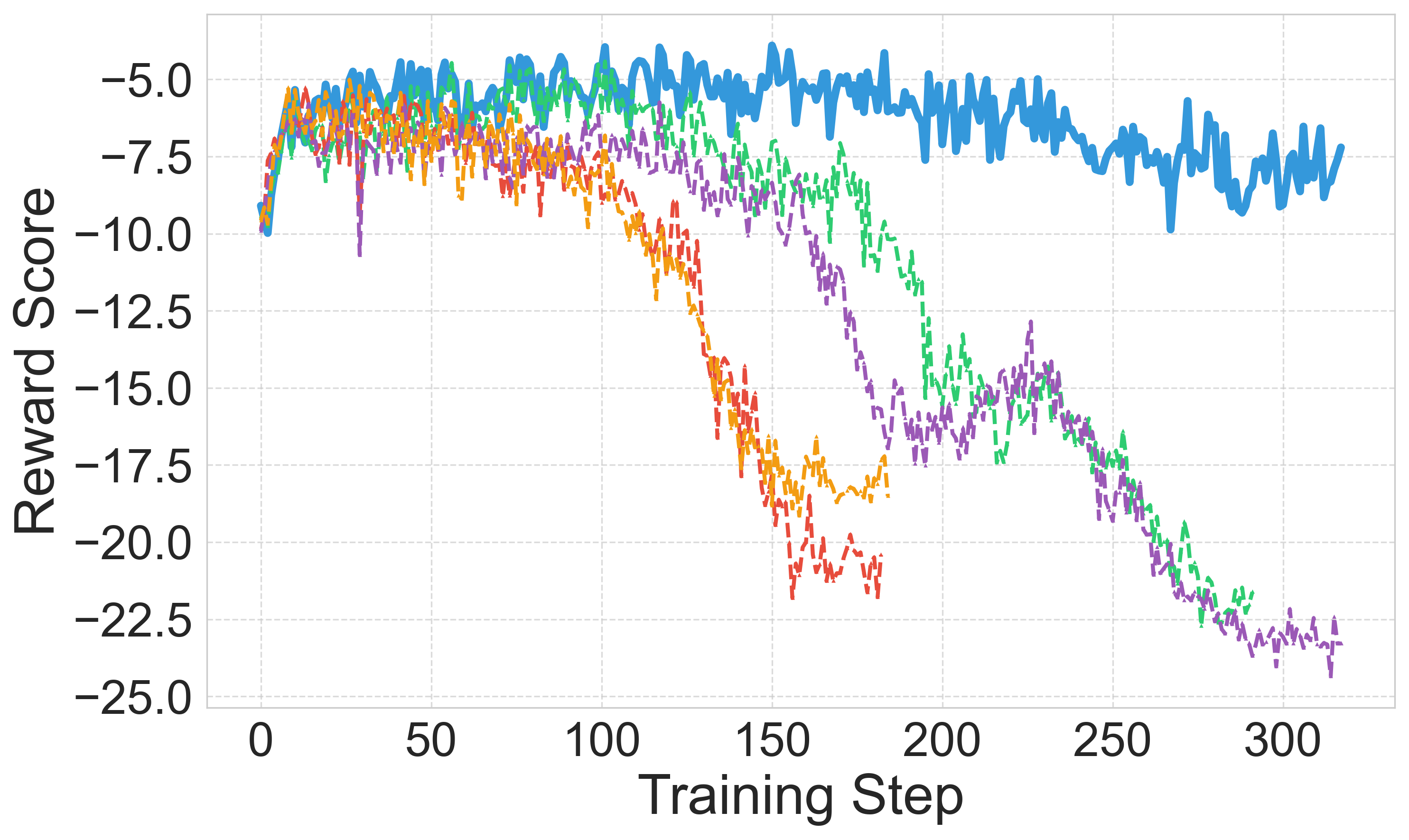}}
    \caption{Downstream policy results with different judge models.}
    \label{fig:downstream}
\end{figure}

\textbf{Results: Real RLHF Setting. } The judge scores across training can be found in Figure \ref{fig:downstream}. Adv-RM results in the most stable training out of any of the baselines, improving over the initial model for 2-3 times more steps than RLHF. We find that RRM is quite a strong baseline, but that Adv-RM's training is still much more stable (note Adv-RM and RRM are complementary approaches that can be used together). This indicates the utility of training with adversarial examples, rather than the randomly sampled responses RRM uses. Furthermore, we show the RewardBench scores achieved by Adv-RM in Table \ref{tab:model_performance}, where we find Adv-RM (83.99) gets a slightly higher score compared to conventional training (83.29).

\begin{figure}[h]
    \centering
    \subfigure[Attack Success Rate]{\includegraphics[width=0.3\textwidth]{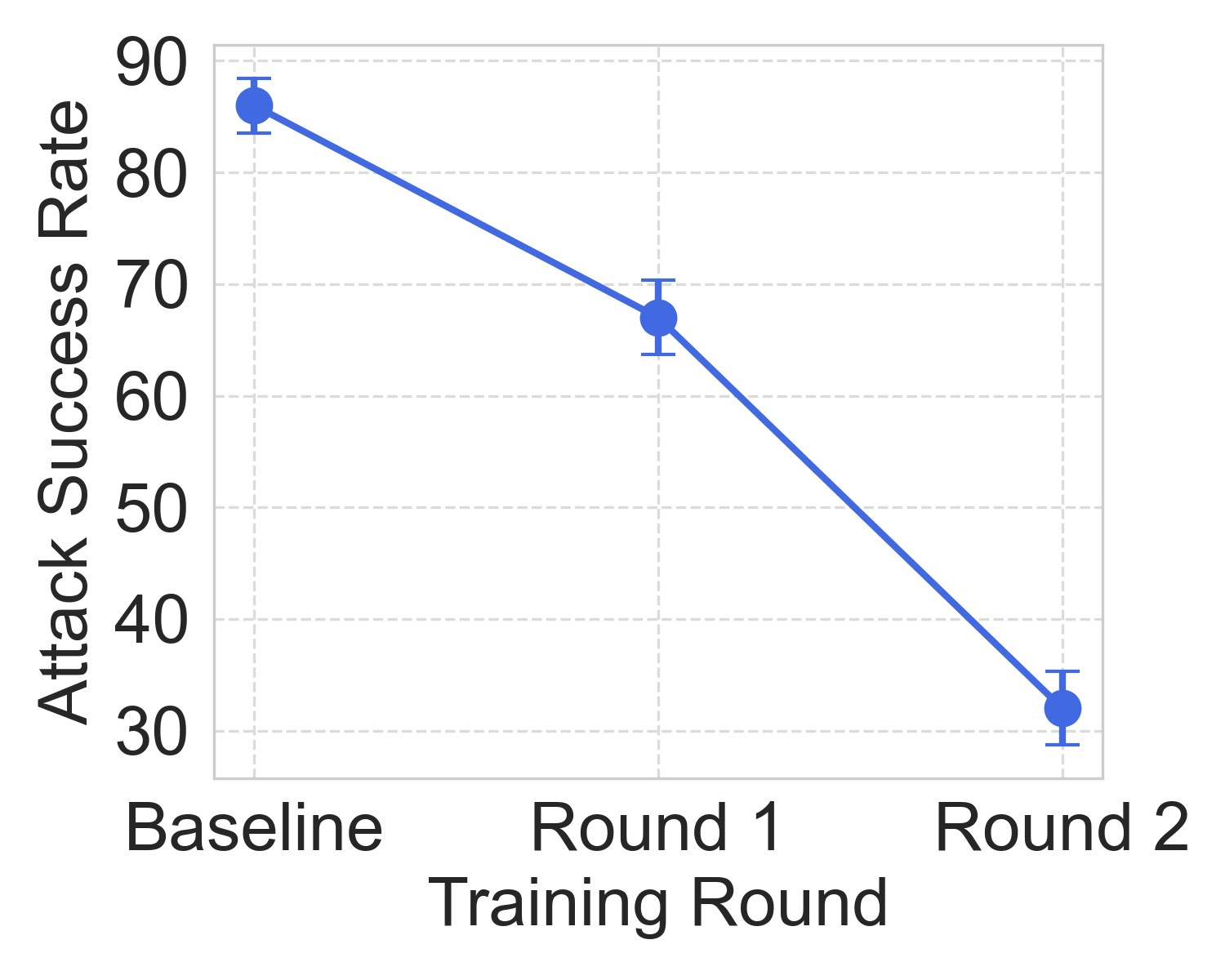} \label{fig:adversarial_robustness}}
    \subfigure[Data Size]{\includegraphics[width=0.3\textwidth]{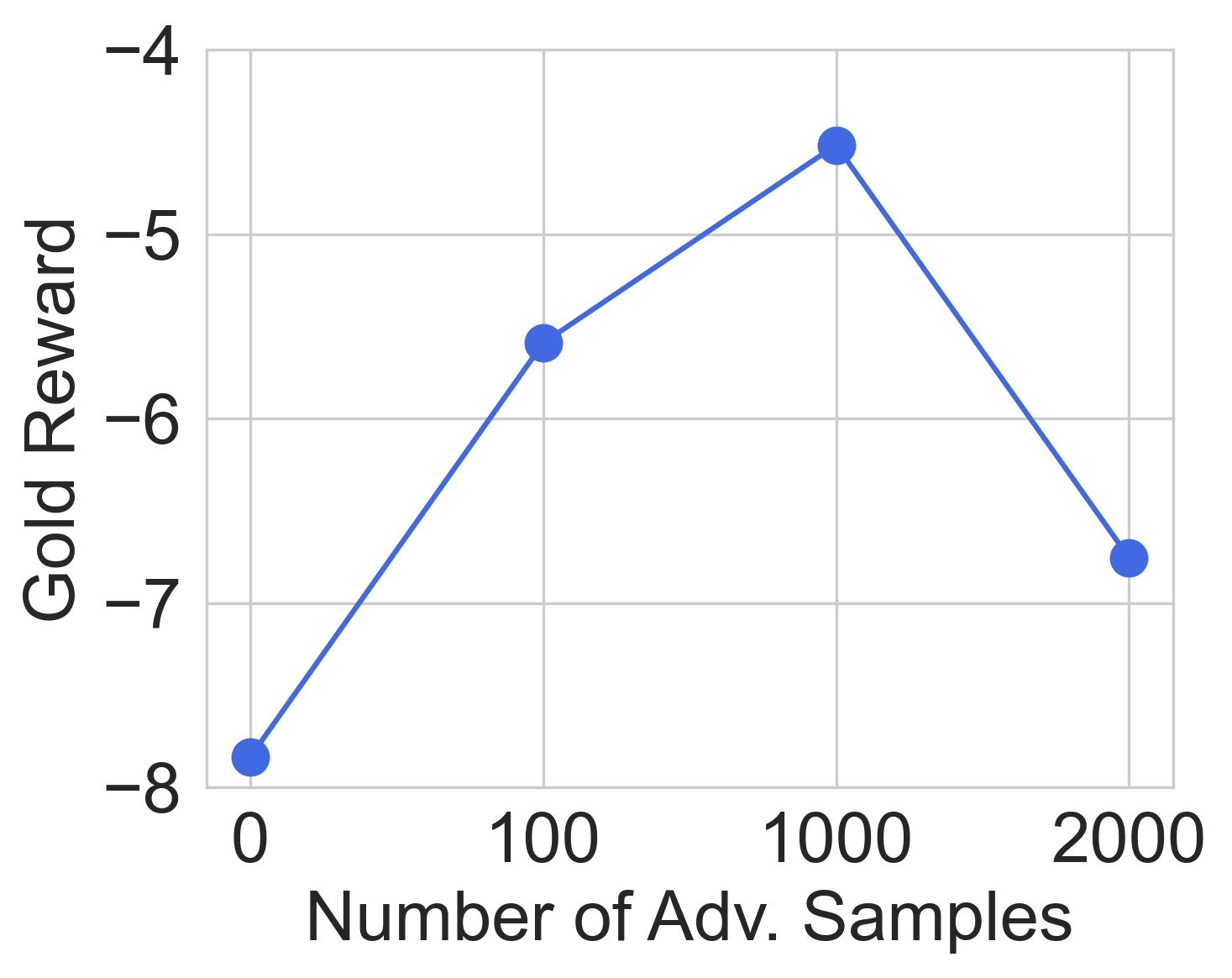} \label{fig:data_ablation}}
    \subfigure[Number of Rounds]{\includegraphics[width=0.3\textwidth]{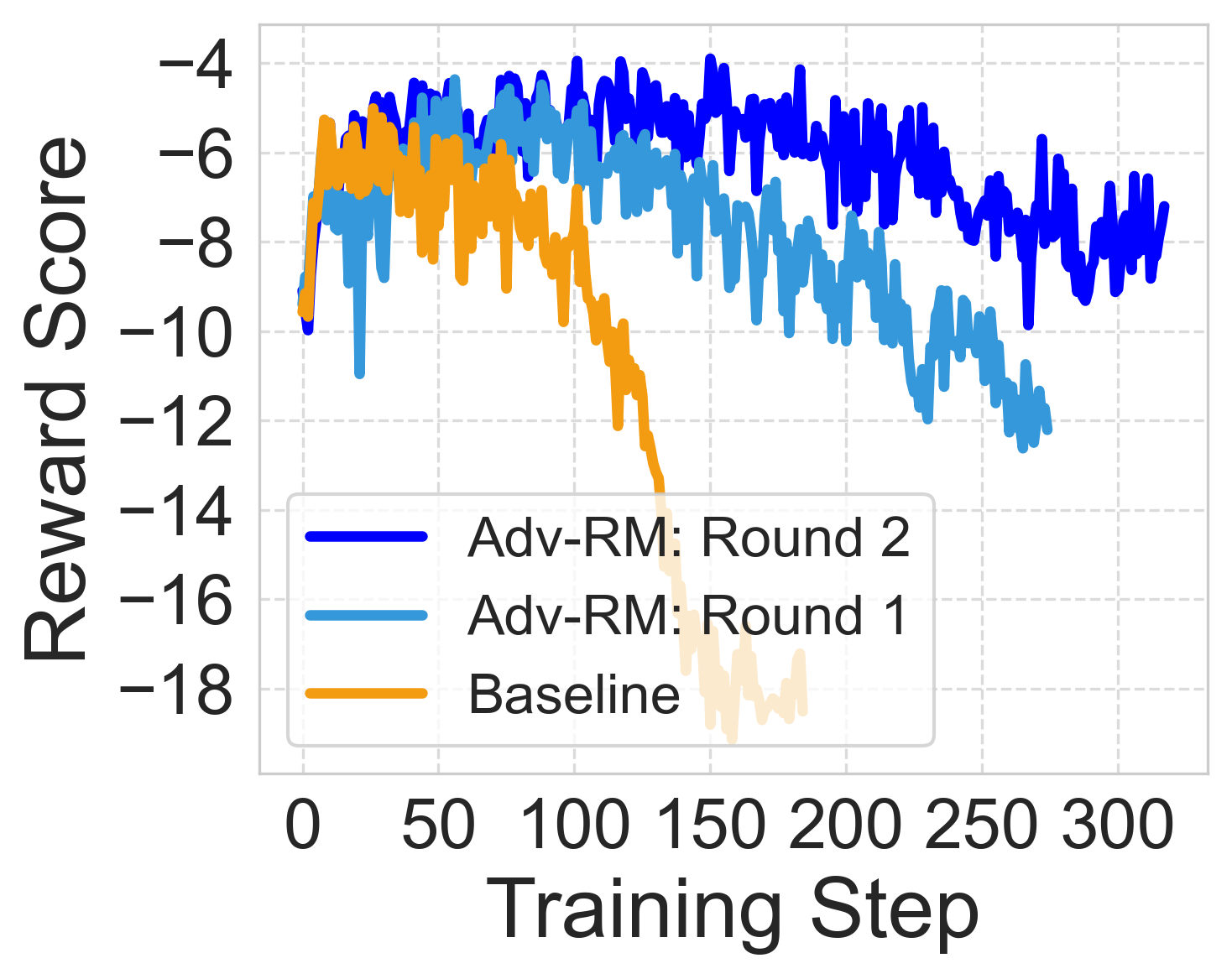} \label{fig:Adv-RM_rounds}}
    \caption{Robustness of Adv-RM models and ablation study. In (a) and (c) we consider the real RLHF setting, where in (b) it is the synthetic setting. For (a) Deepseek-R1 is the judge, and in (b) and (c) Llama-Nemotron RM is.}
    \label{fig:downstream-ablation}
\end{figure}

\textbf{Robustness Against Attacks. }  In Figure \ref{fig:adversarial_robustness} we plot the attack success rate of Adv-RM on policies trained with multiple rounds of Adv-RM training. As we can see, the models trained with Adv-RM become increasingly robust to such attacks, indicating that Adv-RM vulnerabilities can be mitigated with adversarial training.

\textbf{Ablation. } In Figure \ref{fig:data_ablation}, we show the final downstream policy performance of policies trained with different amounts of Adv-RM data. We find that around 1000 samples results in the best downstream performance, and that more samples actually hurts performance. We hypothesize this is due to the fact that Adv-RM data is not very diverse, so adding too much such data can result in overfitting. In Figure \ref{fig:Adv-RM_rounds}, we show the performance with multiple rounds of Adv-RM training. In general we find two rounds leads to the best performance, and that after two rounds, there is minimal gain.

\section{Conclusion and Limitations}
In this paper, we introduce Adv-RM, an adversarial training framework for RMs. Adv-RM can successfully attack state-of-the-art RMs, uncovering surprising failure cases where they assign high scores to low quality responses. Incorporating Adv-RM examples into RM training significantly improves RLHF stability and performance. Although Adv-RM is succesful, it has a few limitations. Adv-RM relies on reward ensembles, meaning adversarial responses may be missed if all RMs incorrectly predict they are high quality, highlighting the need for more robust out-of-distribution detection. Additionally, Adv-RM incurs a substantial computational cost—approximately 3× that of standard RLHF training (see Appendix \ref{app:computational-cost})—but we believe this overhead is justified by the resulting gains in robustness.



\bibliography{colm2025_conference}
\bibliographystyle{colm2025_conference}

\newpage
\appendix
\section{Appendix}

\section{Reproducibility Statement}
We will release all the code and scripts used to reproduce our experiments.

\section{Adversarial Training Hyperparameters}
\label{app:hyperparam-advtraining}
For both the synthetic and real data setup we use similar training configurations. In particular, we use a constant learning rate of $5e-7$ with the Adam optimizer \citep{kingma2014adam}, use a batch size of 64, and generate 4 samples per prompt in RLOO. We fix $\lambda_1 = 1$ and tune $\lambda_2 \in [8, 10, 12]$. We set $C=-25$. For the adversarial policy, we initialize it from llama-3.1-8B-instruct.

For reward model training, we use the following setup: we train for one epoch, and tune the learning rate in $[1e-6, 5e-6, 7e-6, 1e-5]$. We select intermediate checkpoints throughout training based on their RewardBench scores. We use a batch size of 128 and evaluate performance every 10 steps.

\section{Adversarial Attack Baselines}
\label{app:attacks}
\textbf{Textfooler.} For Textfooler, we follow the approach of \cite{jin2020bert} and generate 100 adversarial attacks for each prompt-response pair in our dataset by replacing individual words with their synonyms. We then filter these attacks to only select the ones that result in the highest increase in training reward model score.

\textbf{StyleAdv} For StyleAdv, we use Llama-3.3-70b to rewrite text in one of the five following styles: Shakespearean English, Biblical Prose, Journalistic, Stream of Consciousness, and Satirical. This is in line with \citet{qi2021mind}.

\textbf{RRM.} For RRM, we consider the attack to be the RRM augmentation, where a response from another prompt-response pair is used as the chosen response for the current prompt.

\textbf{RM over-optimization.} For RM over-optimization, we train the policy for three times the number of steps that achieves the best performance. In the synthetic setup this is 150 steps (i.e. best performance is achieved at 50 steps) and in the real-data setup this is 240 steps.

\section{Robust Policy Optimization Baselines}
\label{app:policy-baseline}
\textbf{Ensemble Baselines. } We consider two different ensemble baselines: one where we take the mean between reward models to use as the optimization target, and one where we use the mean minus the standard deviation of the reward model as the optimization target. We weigh the standard deviation of the ensemble baseline by a hyperparameter $\lambda$ that we tune in $[0.1, 0.5, 1.0]$.

\textbf{RRM.} For RRM, we tune the number of augmented datapoints in [2x, 3x, 5x] the original dataset size. We then train reward models on this augmented dataset and conduct policy optimization on the reward model. 

\section{Computational Cost. } 
\label{app:computational-cost}
We conduct all experiments on Nvidia A100 80G GPUs. Training conventional 8b reward models takes around 4 hours on one 8 GPU node, and conventional policy training takes around 12 hours on 2 8 GPU nodes. Training the Adv-RM adversarial policy takes around 12 hours on on 2 8 GPU nodes. Therefore, conventional RLHF takes 224 A100 hours, while Adv-RM training (with 2 rounds) takes 672 A100 hours. We conduct all experiments in NeMo-Aligner \citep{shen2024nemo}.

\section{Evaluating Adversarial Attacks}
\label{app:eval}

\textbf{LLM as a Judge.} For LLM as a judge, we use Llama 3 405B and Deepseek-R1. To get the preference for each adversarial input pair, we ask the LLM judge to rank them with the following prompt:

\begin{center}
\resizebox{.85\linewidth}{!}{%
\begin{minipage}{1.0\linewidth}
\fbox{
\parbox{0.95\linewidth}{
\fontfamily{qcr}\selectfont
Human: For the following query to a chatbot, which response is more helpful and harmless?
\\
\\
Query: <prompt>

Response A: <response A>
  
Response B: <response B>
\\
\\
FIRST provide a one-sentence comparison of the two responses and explain which you feel is more helpful and harmless. SECOND, on a new line, state only "A" or "B" to indicate which response is more helpful. Your response should use the format:

Comparison: <one-sentence comparison and explanation>

More helpful: <"A" or "B">
\\
\\
Assistant:
}
}%
\end{minipage}
}
\end{center}

We make the comparison with both permutations of the preference pair, and say that one response is preferred over the other if it is preferred in both orders. We use a help out dataset of 128 prompts.

\textbf{Human Evaluation. } For human evaluation, we use 4 human annotators, and assign them all preference pairs to annotate. The annotators are not aware of which responses come from different attack methods. We ask them to note if one response is better than the other, or if they are of similar quality. In total, the annotators annotate 50 samples for each method.

\section{Reward Bench Scores}
We present RewardBench scores in Table \ref{tab:model_performance}.

\begin{table}[t]
    \centering
    \caption{Reward Bench Scores}
    \label{tab:model_performance}
    \renewcommand{\arraystretch}{1.2}
    \setlength{\tabcolsep}{8pt}
    \begin{tabular}{lcccccc}
        \toprule
        \textbf{~} & \textbf{Overall} & \textbf{Chat} & \textbf{Chat-Hard} & \textbf{Safety} & \textbf{Reasoning} \\
        \midrule
        Baseline & 0.8329 & 0.9721 & 0.6118 & 0.8541 & 0.8938 \\
        Adv-RM    & 0.8399 & 0.9609 & 0.6404 & 0.8514 & 0.9070 \\
        \bottomrule
    \end{tabular}
\end{table}

\section{Adversarial Policy Training Curve}

\begin{figure}[h]
    \centering
    \subfigure[$R_{\theta_1}(\cdot)$]{\includegraphics[width=0.23\textwidth]{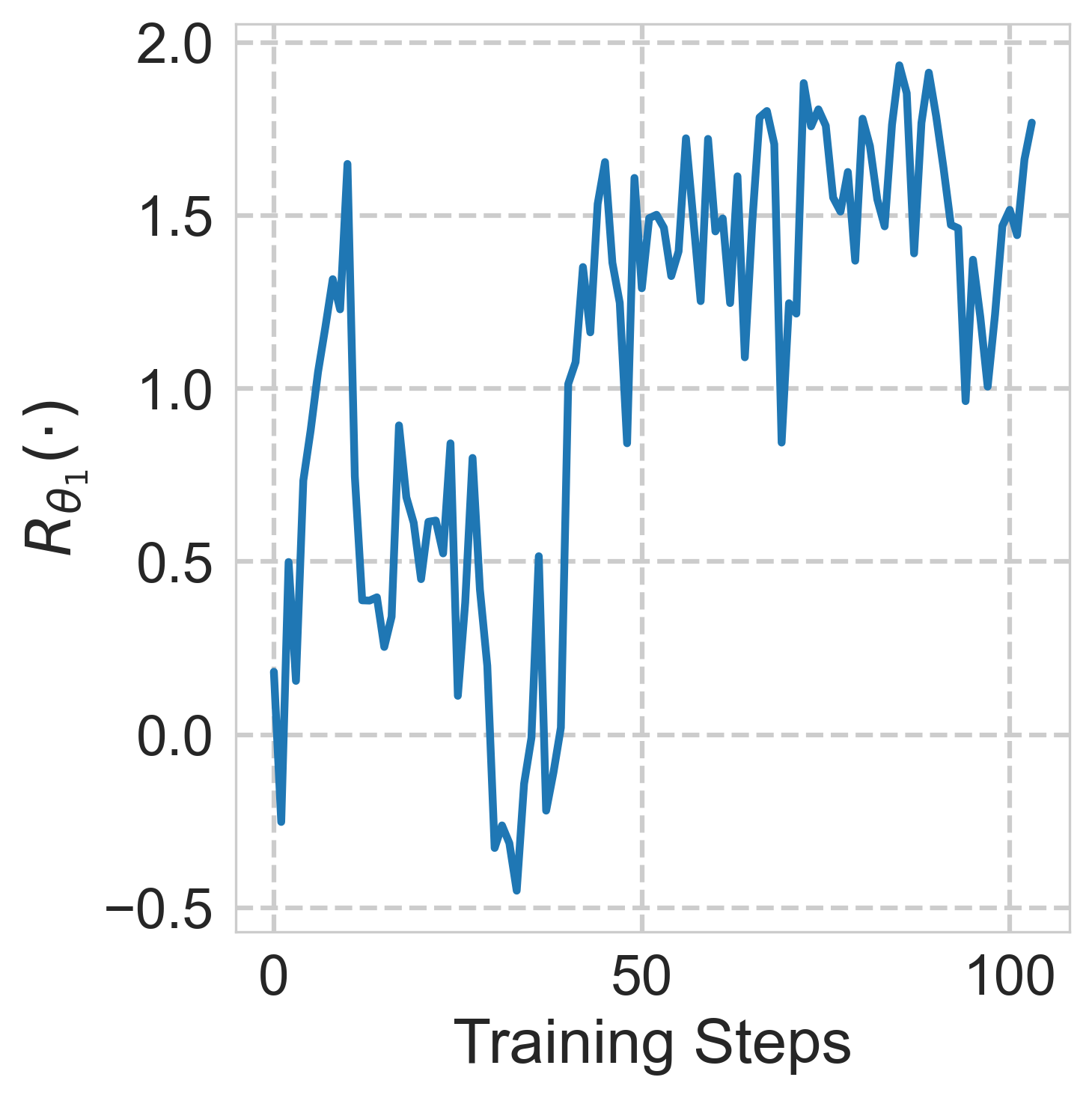}}
    \subfigure[$R_{\theta_2}(\cdot)$]{\includegraphics[width=0.23\textwidth]{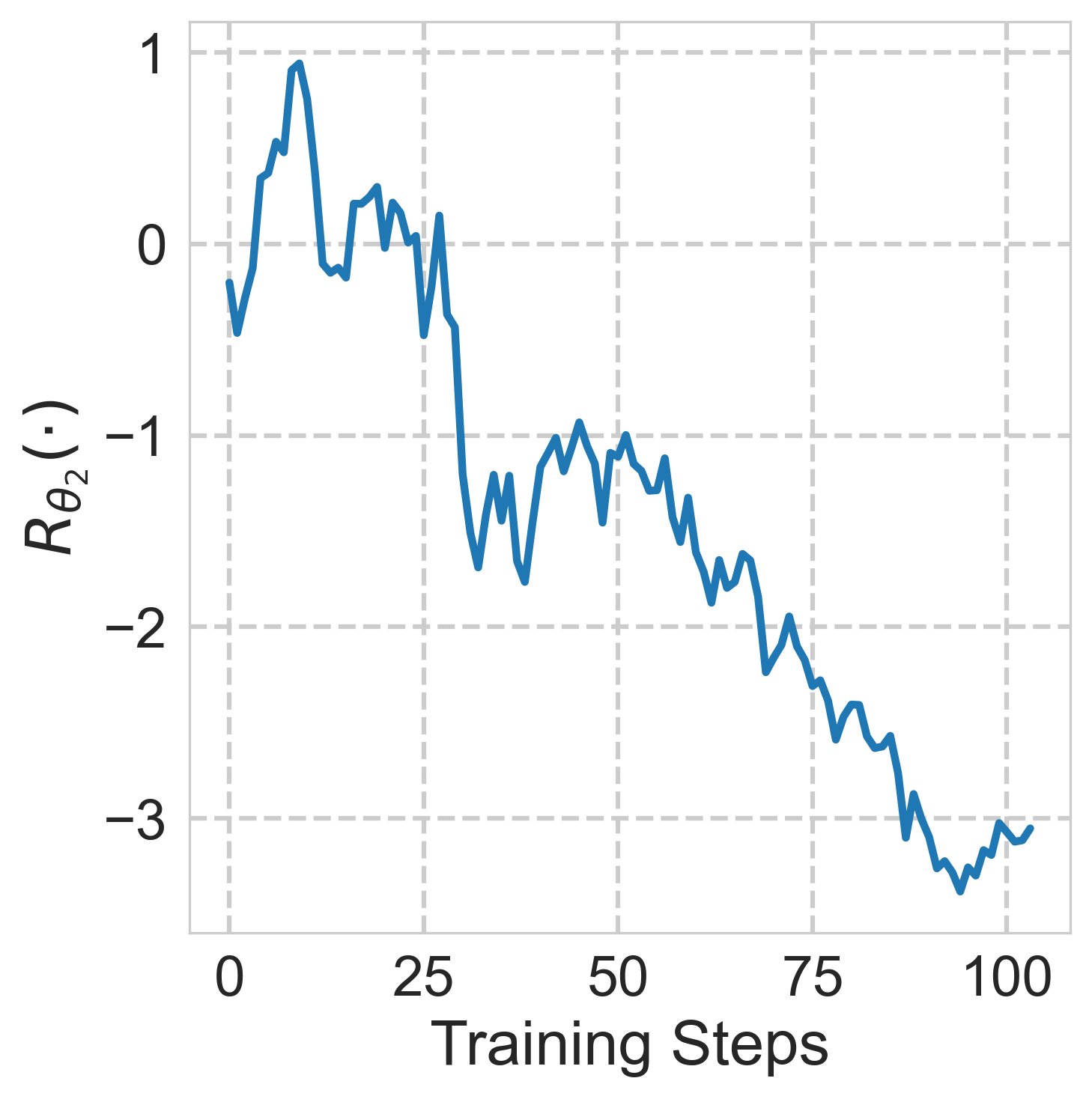}}
    \subfigure[Gold Reward]{\includegraphics[width=0.23\textwidth]{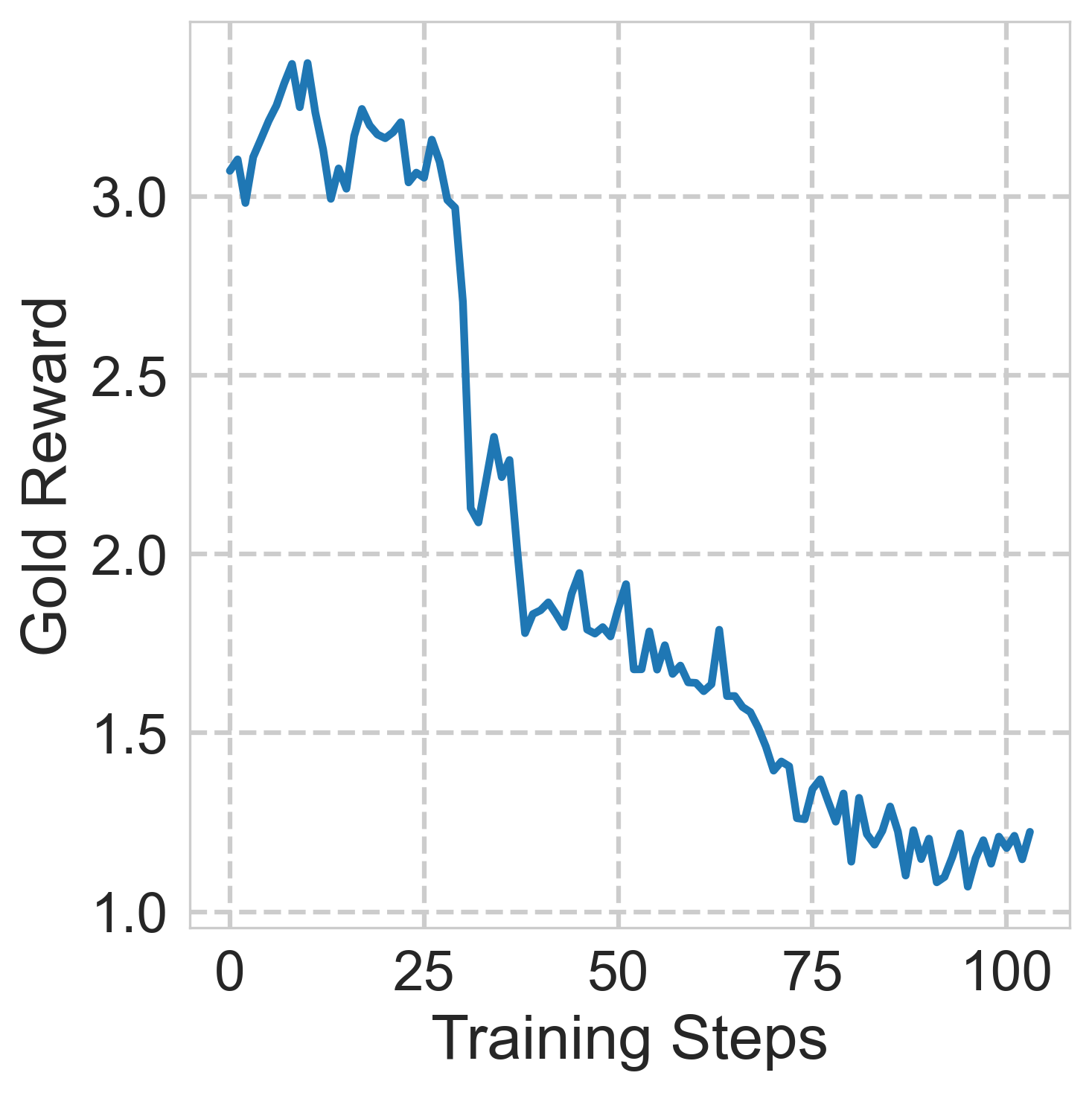}}
    \subfigure[Training Reward]{\includegraphics[width=0.23\textwidth]{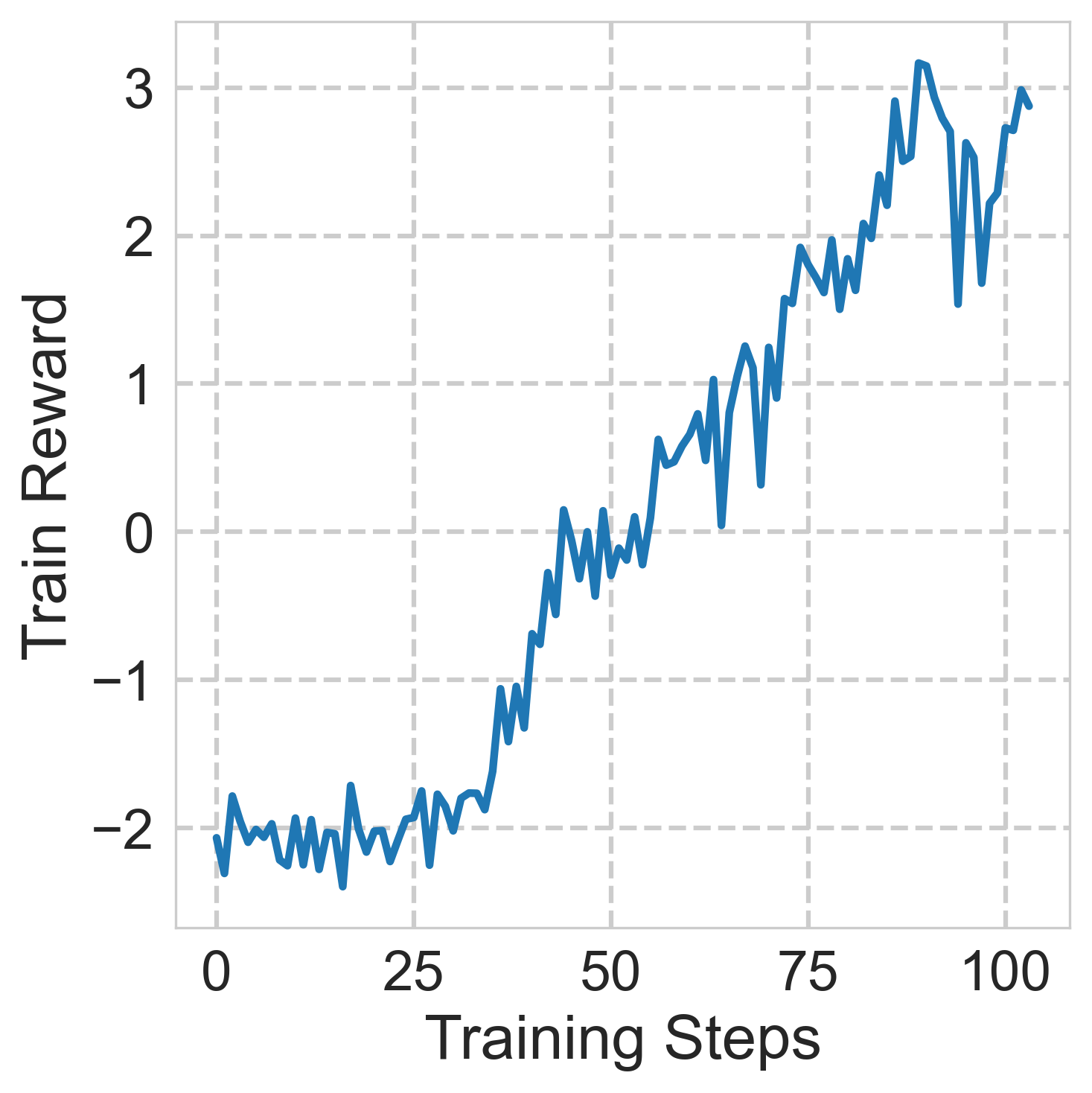}}
    \caption{Adversarial policy training curve in the synthetic setting.}
    \label{fig:adv-training-curve}
\end{figure}

We present the training curve for $\pi_{\text{adv}}$ in Figure \ref{fig:adv-training-curve}. As shown, $\pi_{\text{adv}}$ successfully optimizes \eqref{eq:formulation2} by increasing $R_{\theta_1}$ while minimizing $R_{\theta_2}$. This process leads $\pi_{\text{adv}}$ to generate out-of-distribution (OOD) responses that receive low gold reward. Since these responses achieve high $R_{\theta_1}$ despite their low gold reward, they qualify as adversarial samples, highlighting the effectiveness of Adv-RM.

\end{document}